\documentclass{article}

\usepackage{arxiv}

\usepackage[utf8]{inputenc} % allow utf-8 input
\usepackage[T1]{fontenc}    % use 8-bit T1 fonts
\usepackage{hyperref}       % hyperlinks
\usepackage{url}            % simple URL typesetting
\usepackage{booktabs}       % professional-quality tables
\usepackage{amsfonts}       % blackboard math symbols
\usepackage{nicefrac}       % compact symbols for 1/2, etc.
\usepackage{microtype}      % microtypography
\usepackage{lipsum}		% Can be removed after putting your text content
\usepackage{graphicx}
\usepackage{natbib}
\usepackage{doi}
\usepackage{longtable}
\usepackage{float}
\usepackage{amsmath}

\title{CoLU: \underline{Co}nfidence-Weighted \underline{L}ayerwise \underline{U}nlearning 
for Large Language Models}

\author{
Jingkai Guo \quad Ziyi Zhao \quad Chaitali Chakrabarti \quad Deliang Fan \\
\textit{School of Electrical, Computer and Energy Engineering} \\
\textit{Arizona State University} \\
Tempe, AZ, USA \\
\texttt{\{jingkaig,zzhao206,chaitali,dfan\}@asu.edu}
}

\renewcommand{\shorttitle}{CoLU}

\hypersetup{
pdftitle={A template for the arxiv style},
pdfsubject={q-bio.NC, q-bio.QM},
pdfauthor={Jingkai Guo, Ziyi Zhao, Chaitali Chakrabarti, Deliang Fan},
pdfkeywords={First keyword, Second keyword, More},
}

\begin{document}
\maketitle

\begin{abstract}
Large language model (LLM) unlearning aims to forget certain
information while preserving others, without retraining
from scratch. SoTA methods such as NPO, SimNPO, and SatImp balance forgetting and retain through whole model fine-tuning, while FOM-UL explores selective layer
updates. Prior works show that suppressing target information can damage retained
knowledge. This trade-off can vary across layers even under the same
unlearning loss. In this work, we propose Confidence-Weighted Layerwise Unlearning (CoLU), which combines
a novel confidence-weighted forgetting and retain regularization while updating
a single selected transformer layer, rather than the whole model for unlearning. 
To reduce CoLU compute cost, we further propose Brief Layer Screening (BLS), which uses a newly defined normalized forgetting--retention(NFR) score to first rank layers through a brief unlearning trials using a subset of forgetting dataset and fewer epochs. BLS pre-selects top-k ranked layer candidates, which will then be fully profiled using full data analysis. BLS is proven to reduce the runtime of CoLU by 60.2-79.7\% with almost the same performance. 
On TOFU dataset, CoLU improves both forgetting and retention over NPO, SatImp, and
GA-based FOM-UL in all six TOFU settings, and over SimNPO in five
of the six settings. On MUSE Books and News datasets, CoLU achieves the highest
retained KnowMem and the PrivLeak closest to zero among the compared SOTA
unlearning methods. CoLU presents a new record performance in both unlearning and retain in most settings up to date. Moreover, across the measured settings, average CoLU GPU memory usage is 77.4\% to 79.5\% lower than full-model unlearning.
\end{abstract}

\section{Introduction}
\label{sec:intro}
Large language models can memorize private or sensitive information and
reproduce it in their responses \citep{jang2023knowledge,carlini2021extracting,nasr2025scalable}. Once this information
has been used for training, deleting that data from the training set does not remove its influence on the model \citep{bourtoule2021machine}.
Machine unlearning aims to reduce its influence on the model while
preserving useful capabilities \citep{ NEURIPS2023_062d711f,liu2025rethinking}. Retraining without
the data that is to be forgotten is costly for large models and so methods that modify an existing model using forget and retain data have been proposed \citep{shi2024muse}.

A common approach to unlearning is gradient ascent(GA) on the loss of the
forget data, which lowers the likelihood of targeted responses
\citep{yao2024large}. Although
this can suppress unwanted outputs, continued updates can also damage
performance on retained data \citep{zhang2024npo}. NPO controls the
forgetting process through a negative-preference objective, while SimNPO
removes the reference model and accounts for response length
\citep{zhang2024npo,fan2024simnpo}. These methods aim to improve forgetting while limiting degradation of retained performance, but effective unlearning still requires assessing whether
reduced target likelihood is accompanied by preserved performance on
other inputs. Even after model editing suppresses a target response,
extraction attacks may recover the underlying information
\citep{patil2024defending}.

Token-level reweighting offers finer control over the forgetting signal.
WGA \citep{wang2025wga} downweights low-probability target tokens while SatImp \citep{satimp2025} combines saturation-
and importance-based token reweighting.
The effect of these updates also depends on which parameters are allowed
to change. Prior work has explored selective updates, including
single-layer interventions and constrained optimization of selected LLM
layers \citep{cai2024slug,agrawal2025amnesia}. This motivates us combining
probability-weighted forgetting with a restricted update scope and an
explicit retain objective. Choosing the layer is part of this design:
we seek a location where the unlearning procedure improves forgetting while
preserving retained behavior.

In this paper, we propose \textbf{Co}nfidence-Weighted \textbf{L}ayerwise \textbf{U}nlearning
(CoLU), which combines probability-weighted forgetting and retain regularization while updating a single selected transformer layer.
We use Confidence-Weighted Ascent (CWA), building on the power
reweighting in WGA and SatImp \citep{wang2025wga,satimp2025}, to reduce
each target token's weight as its probability falls. 
% To reduce CoLU compute cost, CoLU uses Brief Layer Screening (BLS) to shortlist candidate layers
% through short unlearning trials, followed by full-data training and final model selection.
Our contributions are as follows:
\begin{itemize}
    \item We propose CoLU, which combines confidence-weighted ascent and
    retain regularization while updating only one selected transformer layer.

    \item To reduce CoLU compute cost, we further propose Brief Layer Screening (BLS), which uses a newly defined normalized forgetting--retention(NFR) score to first rank layers through a brief unlearning trials using a subset of forgetting dataset and fewer epochs. BLS pre-selects top-k ranked layer candidates, which will then be fully profiled using full data analysis. BLS is proven to reduce the runtime of CoLU by 60.2-79.7\% with almost the same performance. 
    
    %\item We define a new normalized forgetting--retention (NFR) score, that combines average relative forgetting gains and normalized retention. It guides both Brief Layer Screening (BLS) candidate selection and final model selection.
    %\item We propose BLS, targeting to reduce compute cost of CoLU, which screens layers using fewer training epochs and a subset of the forgetting dataset to identify top layer candidates for future full data profiling. BLS is proven to reduce the runtime of CoLU by 60.2-79.7\%.
   
    \item On TOFU~\citep{maini2024tofu}, CoLU improves both forgetting and retention over NPO, SatImp, and
    FOM-UL (GA) in all six settings, and over SimNPO in five. On MUSE Books and News~\citep{shi2024muse}, CoLU achieves
    the highest retained KnowMem and the PrivLeak closest to zero among
    the SOTA unlearning methods. To summarize, in this work, CoLU reports a new record performance in both forgetting and retain up to date. 

    \item Ablation studies show that layerwise updates improve NFR for GA,
    NPO, and SatImp on Phi-1.5 Forget-5\%, and that CoLU improves both
    forgetting and retention over full-model SatImp with multiple candidate layers. Across the measured TOFU and MUSE settings,
    single-layer unlearning reduces average GPU memory usage by
    77.4--79.5\% relative to full-model unlearning.
\end{itemize}

\section{Related Work}
\label{sec:related}

\textbf{Machine unlearning.}
Machine unlearning aims to remove the influence of designated training
data while preserving performance on data that should be retained.
Retraining without the designated data provides a reference for this
goal, but its computational cost motivates approximate methods that
modify an existing model. In LLMs, these methods include direct
fine-tuning, localized parameter modification, and approaches that use
auxiliary models \citep{geng2025comprehensive}. A central concern is
whether the intervention suppresses targeted information while preserving
the model's other capabilities.

Among direct optimization methods, gradient ascent (GA) increases the
prediction loss on forget data to reduce the likelihood of targeted
responses. Continued ascent can also degrade retained performance,
motivating objectives that control the strength of forgetting
\citep{zhang2024npo}. Negative Preference Optimization (NPO) uses a
reference-relative sequence objective to slow the utility collapse
associated with GA \citep{zhang2024npo}. SimNPO removes the reference
model and normalizes loss by response length to address reference-model bias
and uneven allocation of unlearning updates \citep{fan2024simnpo}. WGA \citep{wang2025wga} and SatImp \citep{satimp2025} use 
loss reweighting to adjust how much individual tokens contribute to
unlearning on top of the basic GA. Both methods use power re-weighting but weight the token probabilities differently. 

\textbf{Selective parameter updates.}
Selective updates can improve unlearning efficiency and control.
SLUG selects layers using gradient importance and forget--retain alignment,
then tunes the step size along a gradient computed once at the original
model \citep{cai2024slug}. Selective
Amnesia uses causal mediation analysis to guide layer-specific LLM
optimization \citep{agrawal2025amnesia}. SURE combines module-level
gradient saliency with a large learning rate to address the recovery of
forgotten information after quantization \citep{zhang2024sure}.
FOM-UL selects transformer layers using a forget-to-retain significance
score \citep{ranjan2026fomul}. 

% \textbf{Evaluation and robustness.}
% We evaluate on TOFU and both the Books and News domains of MUSE.
% TOFU provides controlled forget and retain subsets based on synthetic
% author profiles \citep{maini2024tofu}. MUSE separately evaluates verbatim
% and knowledge memorization, privacy leakage, retained utility, scalability,
% and sequential unlearning \citep{shi2024muse}.
% \citet{wang2024evaluation} study metric robustness and introduce a
% calibration procedure for comparing methods at restored retain
% performance. We report paired forgetting and retention measurements, with
% selection quality considered separately. 

\section{Method}
\label{sec:method}

\begin{figure}[t]
    \centering
    \includegraphics[width=\linewidth]{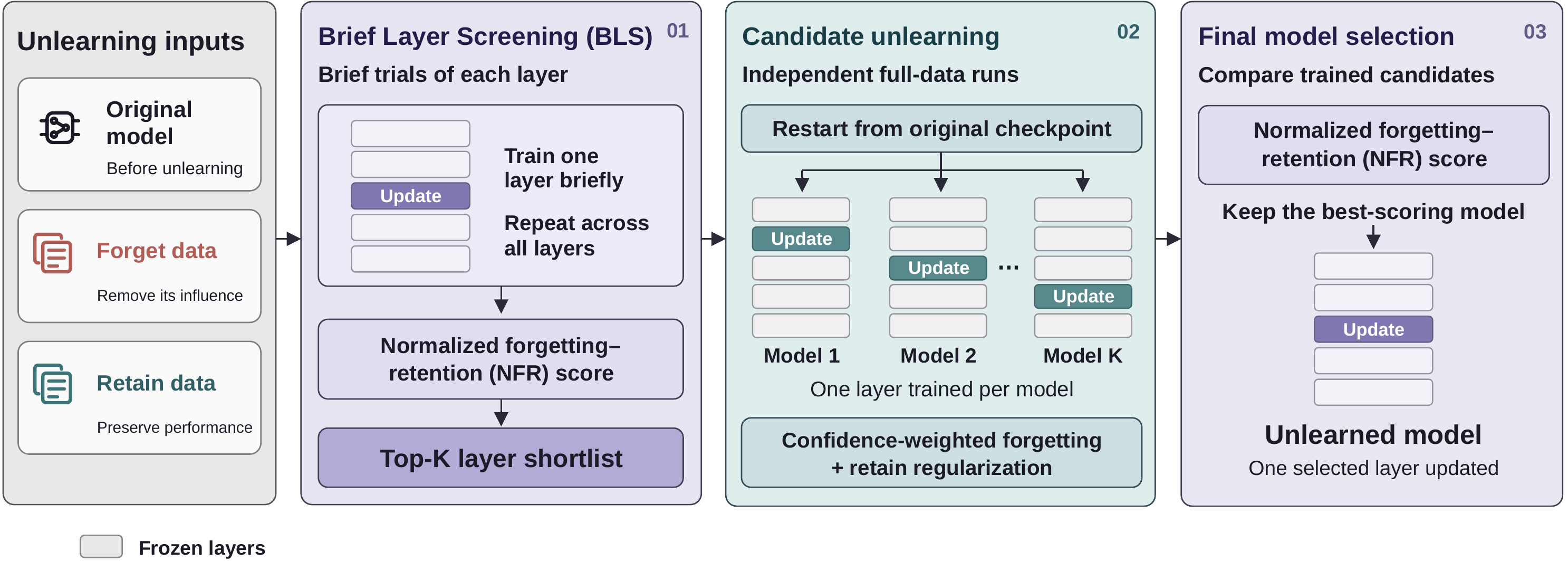}
    \caption{Overview of Confidence-Weighted Layerwise Unlearning (CoLU).
    Brief Layer Screening (BLS) ranks single-layer trials with less epochs using the
    normalized forgetting--retention (NFR) score and keeps the top-$K$ layers.
    Each shortlisted candidate layer is trained independently while keeping the parameters of all other layers frozen. The same score
    selects the final model.}
    \label{fig:overview}
\end{figure}

We propose \emph{Confidence-Weighted Layerwise Unlearning} (CoLU), it rank layers
through a brief unlearning trials using a subset of forgetting dataset and fewer epochs.
BLS pre-selects top-$k$ ranked layer candidates, which will then be fully profiled using full
data analysis. Finally, we select the model
with the highest NFR score.

\textbf{Confidence-Weighted Ascent loss.}
We use a \emph{Confidence-Weighted Ascent} (CWA) loss, a scaled variant adoption
of WGA \citep{wang2025wga} that gives more weight to target tokens
that the model still predicts with high probability.
Let $p_\theta$ be the language model, with
forget set $\mathcal{D}_f$ and retain set $\mathcal{D}_r$. For a target sequence
$y_i$ with context $x_i$, let $p_{i,t}$ denote the probability of token
$y_{i,t}$, which we use as the token's confidence. Let $\mathcal{T}_i$
be the set of valid prediction positions for example $i$, and define
$N_s=\sum_{i\in\mathcal{D}_s}|\mathcal{T}_i|$ for $s\in\{f,r\}$.
These positions follow the causal next-token shift and exclude padding and
masked context tokens.

Following the similar power weighting used in WGA and SatImp
\citep{wang2025wga,satimp2025}, we define the CWA loss on the forget set
$\mathcal{D}_f$ as
\begin{equation}
    \mathcal{L}_f(\theta;\mathcal{D}_f)
    =\frac{1}{N_f}
    \sum_{i\in\mathcal{D}_f}\sum_{t\in\mathcal{T}_i}
    \operatorname{sg}\!\left(Ap_{i,t}^{\alpha}\right)\log p_{i,t},
    \qquad p_{i,t}=p_\theta(y_{i,t}\mid x_i,y_{i,<t}),
    \label{eq:forget-loss}
\end{equation}
where $A>0$ sets the scale and $\alpha>0$ controls how strongly the weights
favor higher-probability tokens. The stop-gradient operator
$\operatorname{sg}$ holds the weights fixed during backpropagation, so
gradients pass only through $\log p_{i,t}$. 
Minimizing the CWA loss performs weighted gradient ascent on the target-token
negative log-likelihood. The loss focuses on confident predictions and reduces the impact of low-confidence ones.
As a token's probability falls, its weight becomes
smaller, reducing its contribution to later updates. A larger $\alpha$ makes
this reduction sharper and also lowers the overall weight magnitude.

To preserve retain-set performance, we use retain regularization
based on cross-entropy on $\mathcal{D}_r$
\citep{zhang2024npo,fan2024simnpo}, yielding the combined loss:
\begin{equation}
    \mathcal{L}(\theta;\mathcal{D}_f,\mathcal{D}_r)
    =\lambda_f\mathcal{L}_f(\theta;\mathcal{D}_f)
    -\lambda_r\frac{1}{N_r}
    \sum_{i\in\mathcal{D}_r}\sum_{t\in\mathcal{T}_i}\log p_{i,t}.
    \label{eq:combined-loss}
\end{equation}
The positive coefficients $\lambda_f$ and $\lambda_r$ set the strength of
the two terms. We minimize the combined objective
over the selected layer's parameters.

\textbf{Layer Update scope.}
We use ``layer'' to represent a complete transformer block with self-attention,
a feed-forward network, normalization, and residual connections.
The original model has parameters $\theta_0$ and $L$ layers indexed from
$0$ to $L-1$. Each run updates all parameters in one candidate layer
$\ell$ and freezes the rest of the model, including the token embeddings
and output head.

\textbf{Brief Layer Screening.}
To limit the cost of evaluating candidate layers, BLS measures how each
responds to a short unlearning run. Every trial starts from $\theta_0$ and uses the same sampled
data and training budget within each experimental setting. On both TOFU
and MUSE, we train on a fixed 20\% subset of the training data for
two epochs. We score the resulting models on fixed forget and
retain screening subsets. And we use same evaluation data for all candidate layers, so the scores are comparable across layers.

We select candidate layers based on both forgetting and retention.
Let $F_\ell^{(m)}$ denote the $m$-th forgetting metric after screening
layer $\ell$, for $m=1,\ldots,M$, and let $R_\ell$ denote the retention
metric. Lower forgetting scores and higher retention scores are preferred.
We define the NFR score as
\begin{equation}
    s_\ell=\frac{100}{2}\left[
        \frac{1}{M}\sum_{m=1}^{M}
        \left(1-\frac{F_\ell^{(m)}}{F_O^{(m)}}\right)
        +\frac{R_\ell}{R_O}\right],
    \label{eq:layer-screening-score}
\end{equation}
where $F_O^{(m)},R_O>0$ are the original model's scores for the same
split. We first average the relative gains across the $M$ forgetting
metrics, then average this forgetting gain with normalized retention.
Thus, forgetting and retention each receive half of the total weight,
regardless of the number of forgetting metrics.
For TOFU, $M=1$ with forget ES and retain ES. For MUSE, $M=2$ with
forget-set VerbMem and KnowMem, and retention is measured by retain-set
KnowMem. We leave the score unclipped, so it can fall outside
the nominal 0--100 range.
We use the same reference scores for every
candidate and return the $K$ highest-scoring layers, breaking ties by
ascending layer index. We set $K=3$ in the screening experiments. 

\textbf{Combined procedure.}
After BLS, we perform candidate unlearning that involves full-data unlearning independently for each of
the $K$ shortlisted candidate layers. Each run restarts from the original checkpoint
and updates only the candidate layer using CWA and retain regularization,
while all other parameters remain fixed. We then evaluate the resulting
models using the same NFR score as in screening and report the results
of the highest-scoring model.

\textbf{Full-screening reference.}
To assess BLS selection, we use full screening (FS) as a reference for
the best observed NFR across layers. FS performs full-data unlearning
separately for each of the $L$ layers. Each run starts from $\theta_0$,
updates one layer using the combined objective in Eq.~\ref{eq:combined-loss},
and freezes all other parameters. We select the resulting model with
the highest NFR score.

\section{Experimental Setup}
\label{sec:experiments}

\textbf{Datasets and models.}
We evaluate our method on TOFU \citep{maini2024tofu} and MUSE
\citep{shi2024muse}. TOFU contains 200 fictitious author profiles, each
described by 20 question--answer pairs. We consider its three standard
forget splits. Our TOFU experiments
use the benchmark-provided Phi-1.5 and Llama-2-7B checkpoints, fine-tuned on
the full TOFU dataset. For MUSE, we evaluate on both Books and News using
the released target checkpoints, based on ICLM-7B and Llama-2-7B,
respectively. Each checkpoint is fine-tuned on the corresponding domain's
forget and retain data.

\textbf{Evaluation metrics.}
For TOFU, our primary metric is extraction strength (ES), following the
robustness analysis of \citet{wang2024evaluation}. ES measures
memorization through the shortest answer prefix needed to recover the
remaining answer. Lower forget-set ES indicates less recoverable target
information, while higher retain-set ES indicates better retention.
We also report TOFU's Forget Quality (FQ) and Model Utility (MU) for
completeness. FQ compares the forget-set truth-ratio distributions of the
unlearned model and a retain-only reference using a Kolmogorov--Smirnov
test. Following \citet{satimp2025}, we distinguish agreement with this
reference from reduced target-information extractability. A low FQ
therefore does not, by itself, indicate failure under our forgetting
objective.
For both MUSE domains, we follow the benchmark's evaluation protocol and report
verbatim memorization (VerbMem) and knowledge memorization (KnowMem) on
the forget set, knowledge memorization on the retain set as a measure of
utility, and privacy leakage (PrivLeak) \citep{shi2024muse}. Lower
forget-set memorization and higher retained utility are preferred.
PrivLeak values closer to zero indicate closer agreement with the
retain-only reference model under the benchmark's membership-inference
evaluation.

\textbf{Baselines.}
We compare CoLU with full-model baselines, including GA,
NPO~\citep{zhang2024npo}, SimNPO~\citep{fan2024simnpo}, and
SatImp~\citep{satimp2025}, on TOFU and MUSE. On TOFU, we also
compare with the multi-layer method FOM-UL~\citep{ranjan2026fomul}.
We evaluate two variants, denoted FU (GA) and FU (CWA) in
Table~\ref{tab:tofu-main}. FU (GA) uses the original paper's GA-based
objective, while FU (CWA) uses CWA loss for forgetting.
We also include the Original and Retrain models as references
for both datasets.

Throughout the tables, Param.\ indicates which parameters are updated:
All denotes the full model, L$j$ denotes transformer layer $j$ (using
zero-based indexing), and Mul denotes progressive updates across
multiple layers. A dash indicates no parameter updates.

\textbf{Implementation and hyperparameters.}
CoLU full-data runs use five epochs on TOFU and eight epochs on MUSE,
for both FS and the BLS shortlisted candidates unlearning.
On both benchmarks, BLS screens each layer for two epochs on a fixed
20\% subset of the training data and selects $K=3$ candidates for
full-data unlearning. BLS uses the same optimizer, batch size,
and loss hyperparameters as the corresponding full-data runs, with a
higher learning rate. We report dataset-specific hyperparameters in
Appendix~\ref{app:hyperparameters}.

\section{Results and Analysis}
\label{sec:results}

% Generated by scripts/build_tofu_main_table.py; edit the generator, not this file.
% Source: ../MyUnlearning/reports/MAIN_RESULTS_COLLECTION.md
% Source SHA-256: ad3d20cce3d9f3515149d9a18c46cba2ef02ab88eb98ff32a997a5f73acb3aa7
% Additional source: tables/tofu_cwa_full_reference.csv (SHA-256: d91935124a6bc4f593b794be9e74eb376dc6b152c8ca68933cd39a47118adc1b)
% Phi Original ES uses the shared references supplied with the 72-layer export:
% Forget ES = 0.6262454641923171 / 0.6990105329441212 / 0.7041712796545074;
% Retain ES = 0.6710792227369352 for each split. Display rounded to 3 decimals.
\begin{table}[t]
\centering
\caption{Comparison of different unlearning methods on TOFU.
$\uparrow$ / $\downarrow$ indicate that larger / smaller values are preferable.
Best ES and MU results are bold; second-best are underlined. FQ is not highlighted as it is less meaningful for unlearning. FU denotes FOM-UL\citep{ranjan2026fomul}.}
\label{tab:tofu-main}
\begingroup
\fontsize{7}{9}\selectfont
\setlength{\tabcolsep}{2pt}
\renewcommand{\arraystretch}{1}
\begin{tabular*}{\textwidth}{@{\extracolsep{\fill}}l*{15}{c}@{}}
\toprule
Method & \multicolumn{5}{c}{Forget-1\%} & \multicolumn{5}{c}{Forget-5\%} & \multicolumn{5}{c}{Forget-10\%} \\
\cmidrule(lr){2-6}\cmidrule(lr){7-11}\cmidrule(l){12-16}
 & Param. & \shortstack{ES\\Un.$\downarrow$} & \shortstack{ES\\Re.$\uparrow$} & FQ$\uparrow$ & MU$\uparrow$ & Param. & \shortstack{ES\\Un.$\downarrow$} & \shortstack{ES\\Re.$\uparrow$} & FQ$\uparrow$ & MU$\uparrow$ & Param. & \shortstack{ES\\Un.$\downarrow$} & \shortstack{ES\\Re.$\uparrow$} & FQ$\uparrow$ & MU$\uparrow$ \\
\midrule
\multicolumn{16}{c}{Phi-1.5} \\
\midrule
Original & \textemdash & 0.626 & 0.671 & 0.003 & 0.502 & \textemdash & 0.699 & 0.671 & 6.5e-13 & 0.502 & \textemdash & 0.704 & 0.671 & 2.2e-16 & 0.502 \\
Retrain & All & 0.041 & 0.756 & 1.000 & 0.498 & All & 0.053 & 0.756 & 1.000 & 0.505 & All & 0.054 & 0.782 & 1.000 & 0.496 \\
\midrule
GA & All & 0.099 & 0.205 & 0.766 & 0.439 & All & 0.147 & 0.293 & 4.4e-8 & 0.455 & All & 0.099 & 0.166 & 4.3e-9 & 0.391 \\
NPO & All & 0.042 & 0.459 & 0.990 & 0.485 & All & 0.058 & 0.476 & 0.866 & 0.486 & All & 0.063 & 0.452 & 0.970 & 0.477 \\
SimNPO & All & 0.224 & \underline{0.616} & 0.003 & 0.504 & All & 0.164 & \underline{0.580} & 2.4e-8 & 0.494 & All & 0.189 & 0.559 & 1.4e-12 & 0.496 \\
SatImp & All & 0.054 & 0.571 & 0.097 & \textbf{0.512} & All & 0.038 & 0.521 & 0.040 & 0.496 & All & 0.035 & 0.458 & 0.007 & \underline{0.509} \\
CWA & All & \textbf{0.016} & 0.518 & 0.579 & \underline{0.506} & All & \underline{0.010} & 0.496 & 0.142 & \underline{0.507} & All & 0.037 & 0.501 & 1.2e-4 & 0.490 \\
FU (GA) & Mul & 0.155 & 0.593 & 0.007 & 0.503 & Mul & 0.249 & 0.448 & 6.6e-12 & 0.478 & Mul & 0.317 & 0.411 & 3.7e-15 & \textbf{0.513} \\
FU (CWA) & Mul & \underline{0.024} & 0.574 & 0.650 & 0.499 & Mul & 0.039 & 0.532 & 0.051 & 0.495 & Mul & 0.012 & 0.485 & 2.2e-4 & 0.478 \\
CoLU (FS) & L20 & 0.036 & \textbf{0.636} & 0.054 & 0.504 & L4 & 0.020 & \textbf{0.586} & 0.088 & \textbf{0.508} & L3 & \underline{0.003} & \textbf{0.579} & 2.6e-9 & 0.501 \\
CoLU (BLS) & L20 & 0.036 & \textbf{0.636} & 0.054 & 0.504 & L23 & \textbf{0.008} & 0.550 & 0.012 & 0.476 & L2 & \textbf{0.002} & \underline{0.560} & 2.2e-6 & 0.493 \\
\midrule
\multicolumn{16}{c}{Llama-2-7B} \\
\midrule
Original & \textemdash & 0.903 & 0.871 & 1.9e-4 & 0.587 & \textemdash & 0.876 & 0.871 & 3.4e-16 & 0.587 & \textemdash & 0.885 & 0.871 & 2.2e-20 & 0.587 \\
Retrain & All & 0.072 & 0.892 & 0.919 & 0.567 & All & 0.066 & 0.897 & 0.965 & 0.575 & All & 0.065 & 0.898 & 0.848 & 0.567 \\
\midrule
GA & All & 0.057 & 0.303 & 0.405 & 0.468 & All & \textbf{0.000} & 0.160 & 3.0e-13 & 0.409 & All & \textbf{0.000} & 0.370 & 1.7e-26 & 0.529 \\
NPO & All & 0.055 & 0.662 & 0.990 & 0.580 & All & 0.047 & 0.673 & 0.713 & 0.590 & All & 0.049 & 0.596 & 0.293 & 0.555 \\
SimNPO & All & 0.132 & 0.746 & 0.014 & \underline{0.598} & All & 0.133 & 0.687 & 0.001 & 0.595 & All & 0.171 & 0.626 & 1.2e-8 & 0.586 \\
SatImp & All & 0.066 & \underline{0.787} & 0.165 & 0.567 & All & 0.006 & 0.709 & 3.1e-12 & \underline{0.602} & All & 0.019 & \underline{0.653} & 5.6e-14 & 0.588 \\
CWA & All & \underline{0.019} & 0.705 & 0.097 & \textbf{0.627} & All & 0.005 & 0.677 & 4.7e-24 & \textbf{0.637} & All & 0.028 & 0.635 & 2.3e-7 & \textbf{0.612} \\
FU (GA) & Mul & 0.089 & 0.559 & 0.014 & 0.516 & Mul & 0.271 & 0.560 & 1.4e-11 & 0.435 & Mul & 0.408 & 0.585 & 1.8e-15 & 0.511 \\
FU (CWA) & Mul & 0.069 & 0.779 & 0.165 & 0.598 & Mul & 0.007 & \underline{0.739} & 0.221 & 0.583 & Mul & 0.005 & 0.602 & 5.1e-17 & \underline{0.604} \\
CoLU (FS) & L4 & \textbf{0.019} & \textbf{0.856} & 0.405 & 0.575 & L3 & \underline{0.003} & \textbf{0.833} & 1.3e-13 & 0.602 & L3 & \underline{0.004} & \textbf{0.849} & 1.8e-15 & 0.600 \\
CoLU (BLS) & L4 & \textbf{0.019} & \textbf{0.856} & 0.405 & 0.575 & L3 & \underline{0.003} & \textbf{0.833} & 1.3e-13 & 0.602 & L3 & \underline{0.004} & \textbf{0.849} & 1.8e-15 & 0.600 \\
\bottomrule
\end{tabular*}
\endgroup
\end{table}

\textbf{Performance on TOFU.}
Table~\ref{tab:tofu-main} compares TOFU results across both models and
all three forget splits. CoLU (BLS) improves both forget and retain ES
over SatImp in all six settings and over SimNPO in five.
On Llama-2-7B Forget-10\%, its retain ES reaches 0.849, dramatically exceeding
SatImp and SimNPO by 19.6 and 22.3 percentage points, respectively,
while its forget ES is 0.004, compared with 0.019 and 0.171.
The exception is Phi-1.5 Forget-5\%, where CoLU (BLS) reduces forget
ES from SimNPO's 0.164 to 0.008 but has lower retain ES
(0.550 versus 0.580).

The full-screening reference, CoLU (FS), achieves the highest retain ES
among the compared unlearning methods in all six settings, together
with lower forget ES than NPO, SimNPO, SatImp, and FU (GA).
CoLU (BLS) selects the same layer and matches this reference in four
settings, including all three Llama-2-7B splits. On Phi-1.5 Forget-5\%
and Forget-10\%, it achieves lower forget ES with lower retain ES
than FS. CoLU (BLS) therefore preserves much of the performance of
full screening while limiting full-data unlearning to the three
BLS-shortlisted candidates.

We include full-model CWA to examine the benefit of layerwise updates.
Full-model CWA achieves lower forget ES than CoLU (BLS) only on Phi-1.5
Forget-1\%, but with much lower retain ES. The consistent retain ES gains
support combining CWA with layer selection to preserve retain-set
behavior during unlearning while achieving similar or lower forget ES
in most settings.

% Generated by scripts/build_muse_main_table.py.
% Source: ../MyUnlearning/reports/MAIN_RESULTS_COLLECTION_MUSE.md
% Source SHA-256: eb9484ff9e334eca45acd3f8f1a4a9791b981d00f205a8f10b3a89e0786b0851
% SimNPO reference: tables/muse_simnpo_reference.csv (user-provided Table 2 image)
% SimNPO reference SHA-256: d2b66c683c9be0a9306ba20cc5bad6e8359b67cc5107607494907ec9c6152d11
% SatImp News reference: tables/muse_satimp_news_reference.csv (user-provided images)
% SatImp News reference SHA-256: 7cab2bd363dc2a6971ddd2b76293e7b7011ab67d5fce4ec7049a8ec2679f442d
% CoLU News reference: reports/MUSE_NEWS_LAYER25_CHECKPOINTS.md
% CoLU News reference SHA-256: 3d34144fe54902daa3ece1f374e7301cd19da52037cb57ffdbfe4e01ec7f19a1
% Books CoLU KnowMem-F: corrected to 0.00 by the user on 2026-09-20.
\begin{table}[t]
\centering
\caption{Unlearning results on MUSE Books and News.
VerbM/KnowM denote VerbMem/KnowMem; F/R denote forget/retain sets.
$\uparrow$ / $\downarrow$ indicate that larger / smaller values are preferable;
$\rightarrow 0$ indicates closing to zero are preferable.
Best values are bold; second-best are underlined.}
\label{tab:muse-main}
\begingroup
\fontsize{8}{10}\selectfont
\setlength{\tabcolsep}{2pt}
\renewcommand{\arraystretch}{1}
\begin{tabular*}{\textwidth}{@{\extracolsep{\fill}}l*{10}{c}@{}}
\toprule
Method & \multicolumn{5}{c}{Books} & \multicolumn{5}{c}{News} \\
\cmidrule(lr){2-6}\cmidrule(l){7-11}
 & Param. & \shortstack{VerbM\\F$\downarrow$} & \shortstack{KnowM\\F$\downarrow$} & \shortstack{KnowM\\R$\uparrow$} & \shortstack{PrivLeak\\$\rightarrow 0$} & Param. & \shortstack{VerbM\\F$\downarrow$} & \shortstack{KnowM\\F$\downarrow$} & \shortstack{KnowM\\R$\uparrow$} & \shortstack{PrivLeak\\$\rightarrow 0$} \\
\midrule
Original & \textemdash & 99.56 & 58.32 & 67.01 & -56.32 & \textemdash & 58.29 & 62.93 & 54.31 & -98.71 \\
Retrain & All & 14.30 & 28.90 & 74.50 & 0.00 & All & 20.75 & 33.32 & 53.79 & 0.00 \\
\midrule
GA & All & \textbf{0.00} & \textbf{0.00} & 0.00 & -24.07 & All & \textbf{0.00} & \textbf{0.00} & 0.00 & \underline{20.14} \\
NPO & All & \textbf{0.00} & \textbf{0.00} & 23.71 & -31.17 & All & 2.53 & 56.93 & 37.58 & 108.91 \\
SimNPO & All & \textbf{0.00} & \textbf{0.00} & \underline{48.27} & \underline{-19.82} & All & 2.34 & 44.84 & \underline{39.65} & 72.93 \\
SatImp & All & \textbf{0.00} & \underline{1.38} & 39.52 & -62.19 & All & \underline{1.93} & \underline{2.75} & 17.62 & -27.31 \\
CoLU (FS) & L28 & \textbf{0.00} & \textbf{0.00} & \textbf{61.85} & \textbf{-19.06} & L25 & 4.63 & 12.81 & \textbf{43.01} & \textbf{-12.28} \\
CoLU (BLS) & L28 & \textbf{0.00} & \textbf{0.00} & \textbf{61.85} & \textbf{-19.06} & L25 & 4.63 & 12.81 & \textbf{43.01} & \textbf{-12.28} \\
\bottomrule
\end{tabular*}
\endgroup
\end{table}

\textbf{Performance on MUSE.}
Table~\ref{tab:muse-main} compares results on MUSE Books and News.
BLS selects the same layer as FS in both domains.
On Books, CoLU records 0.00 on both forget-set memorization metrics
while achieving a retained KnowMem of 61.85. This exceeds SimNPO by
13.58 percentage points and our SatImp reproduction by 22.33 points.
CoLU also achieves the PrivLeak closest to zero among the SOTA methods, at $-19.06$. 

On News, CoLU achieves the highest retained KnowMem at 43.01,
compared with 39.65 for SimNPO and 17.62 for SatImp.
Its forget-set KnowMem is 12.81, well below SimNPO's 44.84, and its
PrivLeak of $-12.28$ is again the closest to zero among the compared
unlearning methods. 

Together, the results on both MUSE domains show that CoLU reduces
forget-set memorization while achieving the highest retained KnowMem
and the PrivLeak closest to zero among the compared unlearning methods.

\section{Ablation Studies}
\subsection{Layerwise Comparison Between Methods}
\label{sec:layerwise-comparison}
% Source: Phi-1.5 Forget-5% layerwise results.
% "CWA (ours)" corresponds to the "Our DEI-test" row.
% Full-model GA/NPO/SatImp/CWA: Phi-1.5 / forget05 in tofu_main_results.json.
% SatImp layer 2: user-supplied exact ES from
% ../../outputs/phi_layerwise/satimp_phi_forget05_beta1_sweep_20260925/satimp_b1_s1_Layer2/ps_exact.json.
% Training time: 97.55 seconds. NFR uses the shared references below.
% Full-model CWA: Forget ES 0.01044 and Retain ES 0.49596,
% also preserved in tofu_cwa_full_reference.csv. The layer-4 CWA run uses LR 2e-4.
% Score = 50 * (1 - Forget ES / 0.6990105329441212
%                   + Retain ES / 0.6710792227369352).
% Shared Phi forget05 references from the supplied 72-layer export.
% All NFR scores are recomputed from the ES values below and rounded once.
\begin{table}[htbp]
\centering
\caption{Layerwise and full-model unlearning on Phi-1.5 with the TOFU
Forget-5\% split. $\uparrow$ / $\downarrow$ indicate that larger / smaller values are preferable.
Layerwise results are picked best from the full sweep of all layers using the NFR score in Eq.~(\ref{eq:layer-screening-score}). The best value in each metric
column is bold.}
\label{tab:phi-forget05-layerwise}
\begingroup
\small
\setlength{\tabcolsep}{3pt}
\renewcommand{\arraystretch}{1}
\begin{tabular*}{\textwidth}{@{\extracolsep{\fill}}l*{7}{c}@{}}
\toprule
Method & \multicolumn{4}{c}{Layerwise} & \multicolumn{3}{c}{Full model} \\
\cmidrule(lr){2-5}\cmidrule(l){6-8}
 & Layer & Forget ES$\downarrow$ & Retain ES$\uparrow$ & NFR$\uparrow$ & Forget ES$\downarrow$ & Retain ES$\uparrow$ & NFR$\uparrow$ \\
\midrule
GA & 0 & 0.201712 & 0.513161 & 73.81 & 0.146975 & 0.293013 & 61.32 \\
NPO & 0 & 0.104086 & 0.593343 & 86.76 & 0.057829 & 0.475640 & 81.30 \\
SatImp & 2 & 0.092705 & \textbf{0.598647} & 87.97 & 0.038465 & \textbf{0.521383} & 86.10 \\
CWA (ours) & 4 & \textbf{0.020209} & 0.586068 & \textbf{92.22} & \textbf{0.010440} & 0.495960 & \textbf{86.21} \\
\bottomrule
\end{tabular*}
\endgroup
\end{table}

To illustrate the performance of layerwise unlearning on different loss functions, Table~\ref{tab:phi-forget05-layerwise} compares layerwise unlearning with
full-model results on Phi-1.5 using the TOFU Forget-5\% split.
For the layerwise runs, we screen all layers and select the best-performing
one for each method using the NFR score to show its upper bound performance. The layerwise configurations
raise NFR from 61.32 to 73.81 for GA, from 81.30 to 86.76 for NPO,
and from 86.10 to 87.97 for SatImp. All three achieve better retention
despite higher forget ES, with GA showing the largest NFR gain.
These results show that selected-layer versions of existing
objectives can improve the forgetting--retention trade-off, although all
three remain below CoLU in NFR score.
Among layerwise results, CoLU achieves the lowest forget ES and
the highest NFR score, while SatImp achieves the highest retain ES.
Full-model CWA achieves lower forget ES than layerwise CWA, while
the layerwise run achieves higher retain ES and raises the NFR score
from 86.21 to 92.22.

\subsection{Full-Screening Results}
\label{sec:full-screen-results}

% Keep this figure on a text page rather than a separate float-only page.
\begin{figure}[htb]
    \centering
    % Both-model vector export; matching source: figures/tofu_all_layers.svg.
    \includegraphics[width=0.95\linewidth]{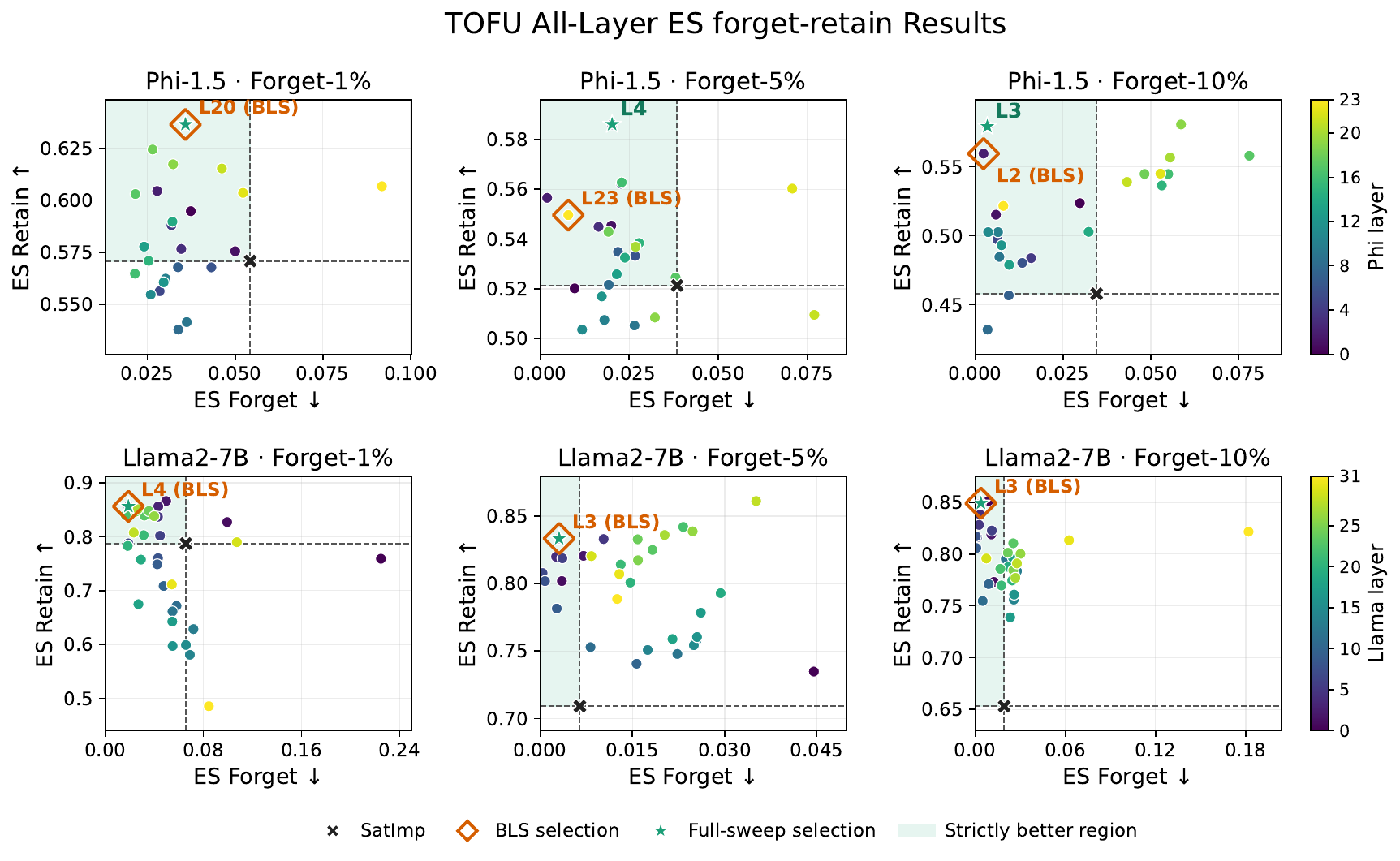}
    \caption{Full-screening results for CoLU on TOFU. Rows show Phi-1.5
    and Llama-2-7B; columns show Forget-1\%, Forget-5\%, and Forget-10\%.
    Each point represents an independently trained layer, colored by layer index.}
    \label{fig:tofu-all-layers}
\end{figure}

Figure~\ref{fig:tofu-all-layers} plots forget ES and retain ES after
applying CoLU independently to all 24 Phi-1.5 layers and all 32
Llama-2-7B layers across the three forget splits. Full-model SatImp
provides the baseline in each setting. The green shaded regions mark
layers with both lower forget ES and higher retain ES than SatImp.
Stars and hollow orange diamonds identify
the FS- and BLS-selected layers reported in Table~\ref{tab:tofu-main},
respectively.

Across the six TOFU settings, 80 of 168 layer runs improve both
metrics over SatImp. These include many layers beyond those selected
by FS and BLS, with effective layers near the beginning and end of
each model. To summarize this pattern, we group layers by relative
depth into early, middle, and late thirds. Pooling the five highest-NFR
layers from each of the six settings gives 30 selections: 17 early
and 13 late, with none in the middle group. Phi-1.5 contributes seven
early and eight late selections, while Llama-2-7B contributes ten early
and five late selections. These results show that CoLU is effective
at different depths, with the strongest results concentrated near
either end of the model. Performance still varies substantially across
layers and forget splits, so selecting among these effective layers
remains important for achieving strong forgetting and retention.

\subsection{Layer Selection Methods}
\label{sec:layer-selection-methods}

To determine how best to select among these layers, we compare BLS with gradient-based selection
scores adapted from SLUG~\citep{cai2024slug},
FOM-UL~\citep{ranjan2026fomul}, and SURE~\citep{zhang2024sure}.
Each selector shortlists three layers for each TOFU setting. We perform
full-data unlearning separately for each candidate and evaluate selection
quality against the corresponding full-screening results.
The gradient scores are computed at the original checkpoint: FOM-UL
ratio compares forget and retain gradient norms, SLUG importance
normalizes the forget gradient norm by the layer's parameter norm,
and SURE-inspired selection uses the forget gradient norm alone.
We apply these scores to complete transformer layers with a fixed
shortlist size. This comparison evaluates their ability to select
layers for CoLU; it does not reproduce the full unlearning procedures
of the original methods.

For each setting, the targets are the three highest-NFR layers in
the full sweep.
\emph{Shortlist hit rate} is the percentage of settings in which a
shortlist contains at least one target. \emph{Recall@3} is the percentage
of the three targets recovered, averaged across settings.
\emph{Winner NFR} is the highest full-data NFR score among the three
shortlisted layers. Hit rate and recall measure recovery of the top
three layers, while winner NFR measures the quality of the selected model.

\begin{table}[htbp]
\centering
\caption{Comparison of layer selection methods on TOFU.
Higher values are better for all three metrics.
Best values per row are bold, including ties. Random-three reports
expected values for uniform sampling of three distinct layers.}
\label{tab:tofu-layer-selection}
\begingroup
\small
\setlength{\tabcolsep}{3pt}
\renewcommand{\arraystretch}{1}
\begin{tabular*}{\textwidth}{@{\extracolsep{\fill}}ll*{5}{c}@{}}
\toprule
Model & Split & \multicolumn{5}{c}{Shortlist hit rate (\%) / Recall@3 (\%) / Winner NFR $\uparrow$} \\
\cmidrule(l){3-7}
 & & \shortstack{BLS\\(ours)} & \shortstack{FOM-UL\\ratio} & \shortstack{SLUG\\importance} & \shortstack{SURE-\\inspired} & \shortstack{Random-three\\(expected)} \\
\midrule
Phi-1.5 & Forget-1\% & \textbf{100}/\textbf{66.7}/\textbf{94.5} & \textbf{100}/\textbf{66.7}/\textbf{94.5} & 0/0/91.3 & 0/0/91.3 & 34.3/12.5/92.4 \\
 & Forget-5\% & \textbf{100}/\textbf{33.3}/\textbf{90.4} & 0/0/89.1 & 0/0/89.4 & 0/0/89.4 & 34.3/12.5/89.8 \\
 & Forget-10\% & \textbf{100}/\textbf{66.7}/91.5 & \textbf{100}/33.3/91.5 & \textbf{100}/33.3/\textbf{92.9} & \textbf{100}/33.3/\textbf{92.9} & 34.3/12.5/88.7 \\
 & Average & \textbf{100}/\textbf{55.6}/\textbf{92.2} & 66.7/33.3/91.7 & 33.3/11.1/91.2 & 33.3/11.1/91.2 & 34.3/12.5/90.3 \\
\midrule
Llama-2-7B & Forget-1\% & \textbf{100}/\textbf{33.3}/\textbf{98.1} & \textbf{100}/\textbf{33.3}/97.3 & 0/0/92.0 & 0/0/92.0 & 26.3/9.4/95.5 \\
 & Forget-5\% & \textbf{100}/\textbf{33.3}/\textbf{97.6} & \textbf{100}/\textbf{33.3}/97.4 & 0/0/95.8 & 0/0/96.0 & 26.3/9.4/96.7 \\
 & Forget-10\% & \textbf{100}/\textbf{66.7}/\textbf{98.5} & 0/0/94.3 & \textbf{100}/33.3/98.3 & 0/0/96.8 & 26.3/9.4/96.3 \\
 & Average & \textbf{100}/\textbf{44.4}/\textbf{98.1} & 66.7/22.2/96.3 & 33.3/11.1/95.4 & 0/0/94.9 & 26.3/9.4/96.1 \\
\midrule
Both models & Overall & \textbf{100}/\textbf{50.0}/\textbf{95.1} & 66.7/27.8/94.0 & 33.3/11.1/93.3 & 16.7/5.6/93.1 & 30.3/10.9/93.2 \\
\bottomrule
\end{tabular*}
\endgroup
\end{table}

Table~\ref{tab:tofu-layer-selection} shows that BLS achieves a 100\%
shortlist hit rate across all six settings, compared with 66.7\% for
FOM-UL ratio, 33.3\% for SLUG importance, and 16.7\% for SURE-inspired
selection. BLS recovers nine of the 18 target positions, giving an
overall Recall@3 of 50.0\%, compared with five for FOM-UL ratio (27.8\%).
It matches or exceeds every comparator on both metrics in each setting.
BLS also achieves the highest or tied-highest winner NFR in five of
six settings and the highest average, at 95.11. The exception is
Phi-1.5 Forget-10\%, where SLUG importance and SURE-inspired selection
obtain higher winner NFR. Compared with Random-three, BLS improves
winner NFR in all six settings, averaging 95.11 versus an expected
93.22, a gain of 1.89 points. For context, the gap between the best and
worst layers in the full sweep averages 12.08 NFR points across the
six settings. Appendix~\ref{app:range-normalized-nfr} reports these
ranges and rescales winner NFR to each setting's observed range.
On this scale, BLS averages 94.8, compared with 77.6 for Random-three.

\subsection{Runtime and Memory Cost}
\label{sec:runtime-memory}

% NPO, SatImp, CoLU (BLS), and BLS runtime ratios.
% FS runtime ratios: CoLU (FS) training time / SatImp training time.
% CoLU (FS): training-only totals.
% Phi-1.5 sums 24 single-layer runs; Llama-2-7B sums 32 single-layer runs.
% Intermediate PS evaluation is excluded from the full-sweep totals.
\begin{table}[htbp]
\centering
\caption{Total training time on TOFU (min).
NPO and SatImp update the full model. CoLU (BLS) includes screening and
training the three selected layers; CoLU (FS) sums training across all layers.}
\label{tab:tofu-runtime-summary}
\begingroup
\small
\setlength{\tabcolsep}{3pt}
\renewcommand{\arraystretch}{1}
\begin{tabular*}{\textwidth}{@{\extracolsep{\fill}}ll*{6}{c}@{}}
\toprule
Model & Split & NPO & SatImp & \shortstack{CoLU\\(BLS)} & \shortstack{Runtime vs.\\SatImp} & \shortstack{CoLU\\(FS)} & \shortstack{Runtime vs.\\SatImp} \\
\midrule
Phi-1.5 & Forget-1\% & 2.69 & 1.48 & 2.90 & $1.96\times$ & 7.29 & $4.93\times$ \\
 & Forget-5\% & 13.09 & 4.62 & 9.89 & $2.14\times$ & 33.88 & $7.33\times$ \\
 & Forget-10\% & 25.54 & 10.60 & 19.05 & $1.80\times$ & 66.93 & $6.31\times$ \\
\midrule
Llama-2-7B & Forget-1\% & 7.18 & 4.83 & 7.68 & $1.59\times$ & 25.01 & $5.18\times$ \\
 & Forget-5\% & 35.65 & 26.28 & 24.09 & $0.92\times$ & 118.44 & $4.51\times$ \\
 & Forget-10\% & 70.78 & 44.18 & 52.86 & $1.20\times$ & 234.05 & $5.30\times$ \\
\bottomrule
\end{tabular*}
\endgroup
\end{table}

Table~\ref{tab:tofu-runtime-summary} compares total training time for
full-model NPO and SatImp with CoLU (FS) and CoLU (BLS).
For Phi-1.5 on the Forget-5\% split, CoLU with BLS takes 9.89~min,
compared with 33.88~min for a full sweep and 4.62~min for full-model
SatImp. The BLS total is $2.14\times$ the SatImp training time.
Averaging the training-time ratio across the three
forget splits gives $1.97\times$ for Phi-1.5 and $1.23\times$ for
Llama-2-7B. Relative to FS, BLS reduces total training time by
60.2--79.7\% across the six settings. These totals include brief
screening and full-data training of the three candidates.

\begin{table}[htbp]
\centering
\caption{GPU memory usage (GiB). TOFU use CWA method as baseline. Single-layer means average
memory usage across runs for all transformer layers.}
\label{tab:unlearning-memory}
\begingroup
\small
\renewcommand{\arraystretch}{1}
\begin{tabular}{lcc}
\toprule
Dataset/model & Full model & \shortstack{Single-layer\\mean} \\
\midrule
TOFU Phi-1.5 & 33.33 & 7.53 \\
TOFU Llama-2-7B & 100.41 & 21.91 \\
MUSE Books & 113.30 & 23.18 \\
MUSE News & 109.30 & 22.78 \\
\bottomrule
\end{tabular}
\endgroup
\end{table}

Speaking of the memory cost, restricting updates to one layer reduces the memory needed for gradients
and optimizer states sine one layer in Llama-2-7B only takes approximately 3\% of the model's parameters. Table~\ref{tab:unlearning-memory} compares full-model
unlearning memory cost with the mean memory usage of single-layer runs
across all transformer layers.
Across the four settings, CoLU reduces average GPU memory usage by 77.4\% to
79.5\%. On TOFU, usage falls from 33.33 to 7.53~GiB for Phi-1.5 and
from 100.41 to 21.91~GiB for Llama-2-7B. On MUSE Books and News,
the single-layer averages are 23.18 and 22.78~GiB, compared with
113.30 and 109.30~GiB for full-model unlearning, respectively.
These savings make CoLU more practical when GPU memory limits the
use of full-model unlearning.

\section{Discussion and Limitations}
\label{sec:discussion}
The experiments above show that CoLU can improve both forgetting and retention over several full-model and multi-layer baselines significantly. BLS reduces the cost of a full sweep, but CoLU can still take longer than full-model unlearning. However, it requires substantially less GPU memory compared to full-model unlearning which make it suitable for many resource-constrained environments. The adapted gradient-based scores from previous work are less effective at identifying promising CoLU layers than BLS, indicating that these criteria do not consistently predict performance after unlearning. Future work should therefore explore selection methods that improve on both BLS and existing criteria, reducing screening cost while reliably identifying effective layers across models and forget splits.

\section{Conclusion}
\label{sec:conclusion}
We presented CoLU, which combines confidence-weighted ascent with
unlearning in a single selected transformer layer. CoLU improves both
forgetting and retention over SOTAs on TOFU and achieves the
highest retained KnowMem and the PrivLeak closest to zero among the
compared unlearning methods on MUSE Books and News. BLS provides results
comparable to full screening on TOFU at substantially lower training
cost. Single-layer updates also reduce GPU memory usage, though
screening remains a source of runtime overhead.

% \vspace{-1em}
\newpage

\bibliographystyle{unsrtnat}
\bibliography{references}

@inproceedings{satimp2025,
author = {Yang, Puning and Wang, Qizhou and Huang, Zhuo and Liu, Tongliang and Zhang, Chengqi and Han, Bo},
title = {Exploring criteria of loss reweighting to enhance LLM unlearning},
year = {2025},
publisher = {JMLR.org},
booktitle = {Proceedings of the 42nd International Conference on Machine Learning},
articleno = {2853},
numpages = {40},
location = {Vancouver, Canada},
series = {ICML'25}
}

@article{maini2024tofu,
  author       = {Pratyush Maini and
                  Zhili Feng and
                  Avi Schwarzschild and
                  Zachary C. Lipton and
                  J. Zico Kolter},
  title        = {{TOFU:} {A} Task of Fictitious Unlearning for LLMs},
  journal      = {CoRR},
  volume       = {abs/2401.06121},
  year         = {2024},
  url          = {https://doi.org/10.48550/arXiv.2401.06121},
  doi          = {10.48550/ARXIV.2401.06121},
  eprinttype   = {arXiv},
  eprint       = {2401.06121},
  bibsource    = {dblp computer science bibliography, https://dblp.org}
}

@inproceedings{shi2024muse,
  author       = {Weijia Shi and
                  Jaechan Lee and
                  Yangsibo Huang and
                  Sadhika Malladi and
                  Jieyu Zhao and
                  Ari Holtzman and
                  Daogao Liu and
                  Luke Zettlemoyer and
                  Noah A. Smith and
                  Chiyuan Zhang},
  title        = {{MUSE:} Machine Unlearning Six-Way Evaluation for Language Models},
  booktitle    = {The Thirteenth International Conference on Learning Representations,
                  {ICLR} 2025, Singapore, April 24-28, 2025},
  publisher    = {OpenReview.net},
  year         = {2025},
  url          = {https://openreview.net/forum?id=TArmA033BU},
  bibsource    = {dblp computer science bibliography, https://dblp.org}
}

@article{zhang2024npo,
    author       = {Ruiqi Zhang and
                  Licong Lin and
                  Yu Bai and
                  Song Mei},
  title        = {Negative Preference Optimization: From Catastrophic Collapse to Effective
                  Unlearning},
  journal      = {CoRR},
  volume       = {abs/2404.05868},
  year         = {2024},
  url          = {https://doi.org/10.48550/arXiv.2404.05868},
  doi          = {10.48550/ARXIV.2404.05868},
  eprinttype   = {arXiv},
  eprint       = {2404.05868},
  bibsource    = {dblp computer science bibliography, https://dblp.org}
}

@inproceedings{fan2024simnpo,
  author       = {Chongyu Fan and
                  Jiancheng Liu and
                  Licong Lin and
                  Jinghan Jia and
                  Ruiqi Zhang and
                  Song Mei and
                  Sijia Liu},
  editor       = {Danielle Belgrave and
                  Cheng Zhang and
                  Laura N. Montoya and
                  Hsuan{-}Tien Lin and
                  Razvan Pascanu and
                  Piotr Koniusz and
                  Marzyeh Ghassemi and
                  Nancy Chen and
                  Iv{\'{a}}n Vladimir Meza Ru{\'{\i}}z and
                  Arturo Loaiza{-}Bonilla},
  title        = {Simplicity Prevails: Rethinking Negative Preference Optimization for
                  {LLM} Unlearning},
  booktitle    = {Advances in Neural Information Processing Systems 38: Annual Conference
                  on Neural Information Processing Systems 2025, NeurIPS 2025, San Diego,
                  CA, USA, December 2-7, 2025 / Mexico City, Mexico, November 30 - December
                  5, 2025},
  year         = {2025},
  url          = {http://papers.nips.cc/paper\_files/paper/2025/hash/02443e4e008231e0af0855f3cc70ed17-Abstract-Conference.html},
  bibsource    = {dblp computer science bibliography, https://dblp.org}
}

@inproceedings{wang2025wga,
  author       = {Qizhou Wang and
                  Jin Peng Zhou and
                  Zhanke Zhou and
                  Saebyeol Shin and
                  Bo Han and
                  Kilian Q. Weinberger},
  title        = {Rethinking {LLM} Unlearning Objectives: {A} Gradient Perspective and
                  Go Beyond},
  booktitle    = {The Thirteenth International Conference on Learning Representations,
                  {ICLR} 2025, Singapore, April 24-28, 2025},
  publisher    = {OpenReview.net},
  year         = {2025},
  url          = {https://openreview.net/forum?id=huo8MqVH6t},
  bibsource    = {dblp computer science bibliography, https://dblp.org}
}

@inproceedings{cai2024slug,
  author       = {Zikui Cai and
                  Yaoteng Tan and
                  M. Salman Asif},
  editor       = {Aarti Singh and
                  Maryam Fazel and
                  Daniel Hsu and
                  Simon Lacoste{-}Julien and
                  Felix Berkenkamp and
                  Tegan Maharaj and
                  Kiri Wagstaff and
                  Jerry Zhu},
  title        = {Targeted Unlearning with Single Layer Unlearning Gradient},
  booktitle    = {Forty-second International Conference on Machine Learning, {ICML}
                  2025, Vancouver, BC, Canada, July 13-19, 2025},
  series       = {Proceedings of Machine Learning Research},
  volume       = {267},
  publisher    = {{PMLR} / OpenReview.net},
  year         = {2025},
  url          = {https://proceedings.mlr.press/v267/cai25f.html},
  bibsource    = {dblp computer science bibliography, https://dblp.org}
}

@article{ranjan2026fomul,
  title = {{Forgetting Only What Matters: Layer-Selective Unlearning toward Robust LLMs}},
  author = {Ranjan, Ravi and Kotevska, Olivera and Polyzou, Agoritsa},
  journal = {arXiv preprint arXiv:2609.10439},
  year = {2026},
  url = {https://arxiv.org/abs/2609.10439}
}

@inproceedings{zhang2024sure,
  author       = {Zhiwei Zhang and
                  Fali Wang and
                  Xiaomin Li and
                  Zongyu Wu and
                  Xianfeng Tang and
                  Hui Liu and
                  Qi He and
                  Wenpeng Yin and
                  Suhang Wang},
  title        = {Catastrophic Failure of {LLM} Unlearning via Quantization},
  booktitle    = {The Thirteenth International Conference on Learning Representations,
                  {ICLR} 2025, Singapore, April 24-28, 2025},
  publisher    = {OpenReview.net},
  year         = {2025},
  url          = {https://openreview.net/forum?id=lHSeDYamnz},
  bibsource    = {dblp computer science bibliography, https://dblp.org}
}

@inproceedings{wang2024evaluation,
  author       = {Qizhou Wang and
                  Bo Han and
                  Puning Yang and
                  Jianing Zhu and
                  Tongliang Liu and
                  Masashi Sugiyama},
  title        = {Towards Effective Evaluations and Comparisons for {LLM} Unlearning
                  Methods},
  booktitle    = {The Thirteenth International Conference on Learning Representations,
                  {ICLR} 2025, Singapore, April 24-28, 2025},
  publisher    = {OpenReview.net},
  year         = {2025},
  url          = {https://openreview.net/forum?id=wUtCieKuQU},
  bibsource    = {dblp computer science bibliography, https://dblp.org}
}

@inproceedings{agrawal2025amnesia,
  title={SHA256 at SemEval-2025 Task 4: Selective Amnesia--Constrained Unlearning for Large Language Models via Knowledge Isolation},
  author={Agrawal, Saransh and Huang, Kuan-Hao},
  booktitle={Proceedings of the 19th International Workshop on Semantic Evaluation (SemEval-2025)},
  pages={2368--2375},
  year={2025}
}

@article{geng2025comprehensive,
  title={A comprehensive survey of machine unlearning techniques for large language models},
  author={Geng, Jiahui and Li, Qing and Woisetschlaeger, Herbert and Chen, Zongxiong and Cai, Fengyu and Wang, Yuxia and Nakov, Preslav and Jacobsen, Hans-Arno and Karray, Fakhri},
  journal={arXiv preprint arXiv:2503.01854},
  year={2025}
}

@inproceedings{jang2023knowledge,
  author       = {Joel Jang and
                  Dongkeun Yoon and
                  Sohee Yang and
                  Sungmin Cha and
                  Moontae Lee and
                  Lajanugen Logeswaran and
                  Minjoon Seo},
  editor       = {Anna Rogers and
                  Jordan L. Boyd{-}Graber and
                  Naoaki Okazaki},
  title        = {Knowledge Unlearning for Mitigating Privacy Risks in Language Models},
  booktitle    = {Proceedings of the 61st Annual Meeting of the Association for Computational
                  Linguistics (Volume 1: Long Papers), {ACL} 2023, Toronto, Canada,
                  July 9-14, 2023},
  pages        = {14389--14408},
  publisher    = {Association for Computational Linguistics},
  year         = {2023},
  url          = {https://doi.org/10.18653/v1/2023.acl-long.805},
  doi          = {10.18653/V1/2023.ACL-LONG.805},
  bibsource    = {dblp computer science bibliography, https://dblp.org}
}

@inproceedings{yao2024large,
 author = {Yao, Yuanshun and Xu, Xiaojun and Liu, Yang},
 booktitle = {Advances in Neural Information Processing Systems},
 doi = {10.52202/079017-3346},
 editor = {A. Globerson and L. Mackey and D. Belgrave and A. Fan and U. Paquet and J. Tomczak and C. Zhang},
 pages = {105425--105475},
 publisher = {Curran Associates, Inc.},
 title = {Large Language Model Unlearning},
 url = {https://proceedings.neurips.cc/paper_files/paper/2024/file/be52acf6bccf4a8c0a90fe2f5cfcead3-Paper-Conference.pdf},
 volume = {37},
 year = {2024}
}

@inproceedings{patil2024defending,
  author       = {Vaidehi Patil and
                  Peter Hase and
                  Mohit Bansal},
  title        = {Can Sensitive Information Be Deleted From LLMs? Objectives for Defending
                  Against Extraction Attacks},
  booktitle    = {The Twelfth International Conference on Learning Representations,
                  {ICLR} 2024, Vienna, Austria, May 7-11, 2024},
  publisher    = {OpenReview.net},
  year         = {2024},
  url          = {https://openreview.net/forum?id=7erlRDoaV8},
  bibsource    = {dblp computer science bibliography, https://dblp.org}
}

@inproceedings{bourtoule2021machine,
title = "Machine unlearning",
author = "Lucas Bourtoule and Varun Chandrasekaran and Choquette-Choo, Christopher A. and Hengrui Jia and Adelin Travers and Baiwu Zhang and David Lie and Nicolas Papernot",
year = "2021",
month = may,
doi = "10.1109/SP40001.2021.00019",
language = "English (US)",
series = "Proceedings - IEEE Symposium on Security and Privacy",
publisher = "Institute of Electrical and Electronics Engineers Inc.",
pages = "141--159",
booktitle = "Proceedings - 2021 IEEE Symposium on Security and Privacy, SP 2021",
address = "United States",
}

@inproceedings{carlini2021extracting,
  title={Extracting training data from large language models},
  author={Carlini, Nicholas and Tramer, Florian and Wallace, Eric and Jagielski, Matthew and Herbert-Voss, Ariel and Lee, Katherine and Roberts, Adam and Brown, Tom and Song, Dawn and Erlingsson, Ulfar and others},
  booktitle={30th USENIX security symposium (USENIX Security 21)},
  pages={2633--2650},
  year={2021}
}

@inproceedings{nasr2025scalable,
  title={Scalable extraction of training data from aligned, production language models},
  author={Nasr, Milad and Rando, Javier and Carlini, Nicholas and Hayase, Jonathan and Jagielski, Matthew and Cooper, A Feder and Ippolito, Daphne and Choquette-Choo, Christopher and Tram{\`e}r, Florian and Lee, Katherine},
  booktitle={International Conference on Learning Representations},
  volume={2025},
  pages={82363--82435},
  year={2025}
}

@inproceedings{NEURIPS2023_062d711f,
 author = {Kurmanji, Meghdad and Triantafillou, Peter and Hayes, Jamie and Triantafillou, Eleni},
 booktitle = {Advances in Neural Information Processing Systems},
 doi = {10.52202/075280-0095},
 editor = {A. Oh and T. Naumann and A. Globerson and K. Saenko and M. Hardt and S. Levine},
 pages = {1957--1987},
 publisher = {Curran Associates, Inc.},
 title = {Towards Unbounded Machine Unlearning},
 url = {https://proceedings.neurips.cc/paper_files/paper/2023/file/062d711fb777322e2152435459e6e9d9-Paper-Conference.pdf},
 volume = {36},
 year = {2023}
}

@article{liu2025rethinking,
  title={Rethinking machine unlearning for large language models},
  author={Liu, Sijia and Yao, Yuanshun and Jia, Jinghan and Casper, Stephen and Baracaldo, Nathalie and Hase, Peter and Yao, Yuguang and Liu, Chris Yuhao and Xu, Xiaojun and Li, Hang and others},
  journal={Nature Machine Intelligence},
  volume={7},
  number={2},
  pages={181--194},
  year={2025},
  publisher={Nature Publishing Group UK London}
}

\clearpage
\appendix

\section{Hyperparameters and Selection Rules}
\label{app:hyperparameters}

\textbf{Shared CoLU settings.}
All CoLU runs use a batch size of 16. On both TOFU and MUSE, BLS trains
each layer for two epochs on a fixed 20\% subset of the training data,
then selects $K=3$ candidates for full-data unlearning. Within each
experimental setting, screening uses the same optimizer, batch size,
and loss hyperparameters as full-data training, with a higher learning
rate. We hold the screening learning rate fixed across all layers.
The full-data learning rates below apply to both FS and the final
candidate runs in BLS.

\textbf{TOFU configurations.}
Full-data runs train one transformer layer for five epochs using paged
AdamW with 32-bit optimizer states and zero weight decay. We set
$A=0.65$, $\lambda_f=1$, and $\lambda_r=0.1$ for all TOFU settings.
Table~\ref{tab:tofu-hyperparameters} lists the learning rates,
$A$ and $\alpha$ values, and loss weights for each model and forget split. Screening and
full-data training use the same $\alpha$ value.

\begin{table}[H]
    \centering
    \caption{CoLU hyperparameters on TOFU. The BLS learning rate applies
    to the brief screening trials.}
    \label{tab:tofu-hyperparameters}
    \begin{tabular}{llcccccc}
        \toprule
        Model & Forget split & Full-data LR & BLS LR & $A$ & $\alpha$ & $\lambda_f$ & $\lambda_r$ \\
        \midrule
        Phi-1.5 & 1\%  & $2\times10^{-4}$ & $4\times10^{-4}$ & 0.65 & 6.0 & 1 & 0.1 \\
                & 5\%  & $2\times10^{-4}$ & $6\times10^{-4}$ & 0.65 & 4.5 & 1 & 0.1 \\
                & 10\% & $2\times10^{-4}$ & $6\times10^{-4}$ & 0.65 & 4.5 & 1 & 0.1 \\
        \midrule
        Llama-2-7B & 1\%  & $1\times10^{-4}$ & $3\times10^{-4}$ & 0.65 & 4.5 & 1 & 0.1 \\
                   & 5\%  & $1\times10^{-4}$ & $4\times10^{-4}$ & 0.65 & 4.5 & 1 & 0.1 \\
                   & 10\% & $1\times10^{-4}$ & $3\times10^{-4}$ & 0.65 & 4.5 & 1 & 0.1 \\
        \bottomrule
    \end{tabular}
\end{table}

\textbf{MUSE configurations.}
On both Books and News, FS trains each layer for eight epochs on the
full data. We select the best checkpoint across epochs, following prior
MUSE evaluations \citep{fan2024simnpo,satimp2025}.
BLS screens layers for two epochs on a fixed 20\% subset,
then restarts each shortlisted candidate from the original checkpoint
and trains it for eight epochs on the full data.
We use a full-data learning rate of $1\times10^{-4}$ and a BLS
screening learning rate of $3\times10^{-4}$ in both domains.
Both domains use AdamW (\texttt{adamw\_torch}) with zero weight decay
and $\lambda_f=1$. We set $\lambda_r=1$ for Books and $\lambda_r=0.4$
for News.
Table~\ref{tab:muse-hyperparameters} summarizes these settings and the
loss hyperparameters, which are shared by screening and full-data runs.

\begin{table}[H]
    \centering
    \caption{CoLU hyperparameters on MUSE. The BLS learning rate applies
    to the two-epoch screening trials.}
    \label{tab:muse-hyperparameters}
    \begin{tabular}{lcccccc}
        \toprule
        Domain & Full-data LR & BLS LR & $A$ & $\alpha$ & $\lambda_f$ & $\lambda_r$ \\
        \midrule
        Books & $1\times10^{-4}$ & $3\times10^{-4}$ & 1 & 4.6 & 1 & 1 \\
        News  & $1\times10^{-4}$ & $3\times10^{-4}$ & 1 & 3.0 & 1 & 0.4 \\
        \bottomrule
    \end{tabular}
\end{table}

\section{Range-Normalized Layer-Selection Scores}
\label{app:range-normalized-nfr}

All Phi-1.5 NFR calculations use the fixed reference
$R_O=0.6710792227$ for all three splits and
$F_O=0.6262454642$, $0.6990105329$, and $0.7041712797$ for
Forget-1\%, Forget-5\%, and Forget-10\%, respectively.

To put winner-score differences in the context of the variation across
layers, we rescale NFR separately for each model and forget split.
For setting $t$, let $s_{\min,t}$ and $s_{\max,t}$ be the lowest and
highest NFR scores in the full layer sweep, and let $s_{a,t}$ be the
winner NFR for selector $a$. We report
\begin{equation}
    z_{a,t}=100\,\frac{s_{a,t}-s_{\min,t}}
                         {s_{\max,t}-s_{\min,t}}.
    \label{eq:range-normalized-nfr}
\end{equation}
A score of 0 corresponds to the worst layer and 100 to the best layer
within that setting. We average $z_{a,t}$ across settings after
rescaling. This supplementary analysis describes relative selection
quality within each sweep; it leaves the NFR selection rule unchanged.
For Random-three, the same transformation applies to the expected
winner NFR because the bounds are fixed within each setting.

% Phi NFR bounds and winners are recomputed from the supplied 72-layer export.
% Phi original retain ES: 0.6710792227369352 for all three splits.
% Phi original forget ES: 0.6262454641923171, 0.6990105329441212,
% 0.7041712796545074 for forget01/05/10, respectively.
% Minima: 87.375044719223, 82.45627239679168, 81.9355131061259,
% 73.164, 89.616, 86.854.
% Maxima: 94.54728783508838, 92.22053651865643, 92.91589006554788,
% 98.0865, 97.638, 98.4955.
% These Phi references are shared with the objective ablation.
% Llama scores retain the supplied historical C/2 values.
% Llama forget05 maximum: 97.638 (layer 3).
% FOM-UL Llama forget05 = 196.292/2 - 50*(0.03511-0.02211)/0.876.
% Normalize per setting, average before rounding, display one decimal.
\begin{table}[H]
\centering
\caption{Observed layer NFR bounds for the supplementary comparison.}
\label{tab:tofu-nfr-ranges}
\begin{tabular*}{\textwidth}{@{\extracolsep{\fill}}llrrr@{}}
\toprule
Model & Split & Minimum (layer) & Maximum (layer) & Range width \\
\midrule
Phi-1.5 & Forget-1\% & 87.38 (L8) & 94.55 (L20) & 7.17 \\
 & Forget-5\% & 82.46 (L21) & 92.22 (L4) & 9.76 \\
 & Forget-10\% & 81.94 (L8) & 92.92 (L3) & 10.98 \\
\midrule
Llama-2-7B & Forget-1\% & 73.16 (L31) & 98.09 (L4) & 24.92 \\
 & Forget-5\% & 89.62 (L0) & 97.64 (L3) & 8.02 \\
 & Forget-10\% & 86.85 (L31) & 98.50 (L3) & 11.64 \\
\midrule
Both models & Average & & & 12.08 \\
\bottomrule
\end{tabular*}
\end{table}

\begin{table}[H]
\centering
\caption{Range-normalized winner NFR ($\uparrow$) from
Eq.~(\ref{eq:range-normalized-nfr}). Best values per row are bold,
including ties.}
\label{tab:tofu-layer-selection-normalized}
\begin{tabular*}{\textwidth}{@{\extracolsep{\fill}}ll*{5}{c}@{}}
\toprule
Model & Split & \shortstack{BLS\\(ours)} & \shortstack{FOM-UL\\ratio} & \shortstack{SLUG\\importance} & \shortstack{SURE-\\inspired} & \shortstack{Random-three\\(expected)} \\
\midrule
Phi-1.5 & Forget-1\% & \textbf{100.0} & \textbf{100.0} & 55.1 & 55.1 & 70.5 \\
 & Forget-5\% & \textbf{81.2} & 67.8 & 71.4 & 71.4 & 75.1 \\
 & Forget-10\% & 87.3 & 87.3 & \textbf{100.0} & \textbf{100.0} & 61.2 \\
 & Average & \textbf{89.5} & 85.0 & 75.5 & 75.5 & 69.0 \\
\midrule
Llama-2-7B & Forget-1\% & \textbf{100.0} & 96.7 & 75.4 & 75.4 & 89.6 \\
 & Forget-5\% & \textbf{100.0} & 97.1 & 77.1 & 79.1 & 88.1 \\
 & Forget-10\% & \textbf{100.0} & 63.8 & 98.6 & 85.1 & 80.9 \\
 & Average & \textbf{100.0} & 85.8 & 83.7 & 79.8 & 86.2 \\
\midrule
Both models & Overall & \textbf{94.8} & 85.4 & 79.6 & 77.7 & 77.6 \\
\bottomrule
\end{tabular*}
\end{table}

Under the bounds in Table~\ref{tab:tofu-nfr-ranges}, layer NFR ranges
span 7.17--24.92 points, with a mean width of 12.08 points.
Table~\ref{tab:tofu-layer-selection-normalized} places each selector's
winner within these ranges. BLS averages 94.8, compared with 77.6 for
Random-three; these are positions on a rescaled score, rather than
percentages of information forgotten.

\section{Effect of Retain Regularization}

\begin{table}[ht]
\centering
\caption{Retain-regularization ablation for CoLU (FS) on
Phi-1.5 across TOFU splits. We report the result with the highest
NFR score from each 24-layer sweep, without a
baseline-dominance filter.}
\label{tab:retain-regularization-ablation}
\begin{tabular}{llccc}
\toprule
Method & Split & Layer & ES Forget $\downarrow$ & ES Retain $\uparrow$ \\
\midrule
\multicolumn{5}{c}{With retain regularization } \\
\midrule
CoLU (FS) & Forget-1\%  & 20 & 0.035851 & 0.636313 \\
CoLU (FS) & Forget-5\%  & 4  & 0.020209 & 0.586068 \\
CoLU (FS) & Forget-10\% & 3  & 0.003480 & 0.579316 \\
\midrule
\multicolumn{5}{c}{Without retain regularization} \\
\midrule
CoLU (FS) & Forget-1\%  & 20 & 0.024284 & 0.458252 \\
CoLU (FS) & Forget-5\%  & $1^{\dagger}$ & 0.000000 & 0.000000 \\
CoLU (FS) & Forget-10\% & 0  & 0.000000 & 0.000192 \\
\bottomrule
\end{tabular}

\smallskip
\begin{minipage}{\linewidth}
\footnotesize
$^{\dagger}$Layers 1--23 tie with zero Forget and Retain ES;
layer 1 is reported by the layer-index tie-break rule.
This tie reflects retention collapse, not successful unlearning.
\end{minipage}
\end{table}

At the existing learning rates and training budgets, removing
retain regularization substantially reduces retention.
On Forget-5\% and Forget-10\%, even full layer search cannot
recover a useful forget--retain trade-off. These results
support the importance of retain regularization in the
current CoLU configuration.

% \clearpage
\section{Variation Across Training Seeds}
\label{app:phi-three-seed}

We evaluate Phi-1.5 with training seeds 1, 2, and 3 on all three TOFU
forget splits, using the configurations in
Appendix~\ref{app:hyperparameters}. Each seed includes a full-data run
for every layer and a separate set of BLS screening trials. FS selects
the highest-NFR result among all 24 layers; BLS selects the highest-NFR
result within its three-layer screening shortlist. For this analysis,
BLS uses the corresponding full-sweep checkpoints, so FS and BLS share
the same evaluation when they select the same layer. The screening
subsets remain fixed across training seeds.

Table~\ref{tab:phi-three-seed-selected} reports the mean and sample
standard deviation of the three selected outcomes for each method and
split. Table~\ref{tab:phi-three-seed-layers} instead summarizes each
fixed layer across seeds. FQ and MU are reevaluated under a common
protocol for all three seeds; these evaluations differ from the
historical evaluations in Table~\ref{tab:tofu-main}. Neither FQ nor MU
is used for selection. FQ means and standard deviations are descriptive
summaries of the individual test $p$-values, not a pooled test.

Seed 1 reuses historical training runs and ES measurements, while
seeds 2 and 3 use the multi-seed training launcher. The observed
dispersion may therefore include execution differences in addition
to training-seed variation. It does not measure variation from
resampling the screening subsets. ES is used for both selection and
reporting, so the selected scores are not independent held-out estimates.

% Source: author-supplied Phi-1.5 consolidated three-seed report (2026-09-25).
% Source SHA-256: 7c4efb76e816c427b963d7bed4eef5c441d61578e1dc825159b28657bdc31742
% Means and sample SDs use seeds 1, 2, 3 (ddof=1). NFR uses supplied scores.
% Both tables aggregate the same supplied per-seed records before display rounding.

\begin{table}[H]
\centering
\caption{Selected CoLU results on Phi-1.5 across three training seeds.
Each metric cell gives the mean above and $\pm$ sample standard deviation
below ($n=3$). Layer indices are zero-based and listed in seed order 1/2/3.
FS selects the highest-NFR layer in each full sweep; BLS selects the
highest-NFR layer within its screening shortlist.}
\label{tab:phi-three-seed-selected}
\begingroup
\small
\setlength{\tabcolsep}{3pt}
\renewcommand{\arraystretch}{1.1}
\begin{tabular*}{\textwidth}{@{\extracolsep{\fill}}llccccc@{}}
\toprule
Method & \shortstack{Layers\\(1/2/3)} & \shortstack{Forget\\ES $\downarrow$} & \shortstack{Retain\\ES $\uparrow$} & NFR $\uparrow$ & FQ $\uparrow$ & MU $\uparrow$ \\
\midrule
\multicolumn{7}{l}{\textit{Forget-1\%}} \\
CoLU (FS) & 20/20/20 & \shortstack{$0.0337$\\$\pm\,0.0019$} & \shortstack{$0.6406$\\$\pm\,0.0048$} & \shortstack{$95.04$\\$\pm\,0.47$} & \shortstack{$0.2079$\\$\pm\,0.1791$} & \shortstack{$0.5035$\\$\pm\,0.0021$} \\
CoLU (BLS) & 20/20/20 & \shortstack{$0.0337$\\$\pm\,0.0019$} & \shortstack{$0.6406$\\$\pm\,0.0048$} & \shortstack{$95.04$\\$\pm\,0.47$} & \shortstack{$0.2079$\\$\pm\,0.1791$} & \shortstack{$0.5035$\\$\pm\,0.0021$} \\
\midrule
\multicolumn{7}{l}{\textit{Forget-5\%}} \\
CoLU (FS) & 4/1/2 & \shortstack{$0.0115$\\$\pm\,0.0101$} & \shortstack{$0.5712$\\$\pm\,0.0202$} & \shortstack{$91.74$\\$\pm\,0.80$} & \shortstack{$0.0152$\\$\pm\,0.0213$} & \shortstack{$0.5030$\\$\pm\,0.0058$} \\
CoLU (BLS) & 23/18/2 & \shortstack{$0.0097$\\$\pm\,0.0102$} & \shortstack{$0.5473$\\$\pm\,0.0029$} & \shortstack{$90.08$\\$\pm\,0.92$} & \shortstack{$0.1167$\\$\pm\,0.1834$} & \shortstack{$0.4951$\\$\pm\,0.0228$} \\
\midrule
\multicolumn{7}{l}{\textit{Forget-10\%}} \\
CoLU (FS) & 3/1/2 & \shortstack{$0.0035$\\$\pm\,0.0011$} & \shortstack{$0.5741$\\$\pm\,0.0095$} & \shortstack{$92.53$\\$\pm\,0.78$} & \shortstack{$2.97\!\times\!10^{-8}$\\$\pm\,5.08\!\times\!10^{-8}$} & \shortstack{$0.5025$\\$\pm\,0.0031$} \\
CoLU (BLS) & 2/2/2 & \shortstack{$0.0023$\\$\pm\,0.0022$} & \shortstack{$0.5614$\\$\pm\,0.0017$} & \shortstack{$91.66$\\$\pm\,0.16$} & \shortstack{$4.96\!\times\!10^{-7}$\\$\pm\,7.85\!\times\!10^{-7}$} & \shortstack{$0.4977$\\$\pm\,0.0071$} \\
\bottomrule
\end{tabular*}
\endgroup
\end{table}

FS and BLS select layer 20 in all three Forget-1\% runs. The selected
layers vary on Forget-5\%, while BLS selects layer 2 in all three
Forget-10\% runs. On Forget-5\% and Forget-10\%, BLS has lower mean
forget ES than FS, accompanied by lower mean retain ES and NFR.

\clearpage
% Source: author-supplied Phi-1.5 consolidated three-seed report (2026-09-25).
% Source SHA-256: 7c4efb76e816c427b963d7bed4eef5c441d61578e1dc825159b28657bdc31742
% Means and sample SDs use seeds 1, 2, 3 (ddof=1). NFR uses supplied scores.
% Both tables aggregate the same supplied per-seed records before display rounding.

% Layerwise statistics computed from the supplied per-seed rows before display rounding.

\begingroup
\small
\setlength{\tabcolsep}{2pt}
\renewcommand{\arraystretch}{1.15}
\begin{longtable}{@{\extracolsep{\fill}}lccccc@{}}
\caption{Full-data CoLU results for every Phi-1.5 layer across three
training seeds. Values are mean $\pm$ sample standard deviation
($n=3$), computed for the same layer across
seeds. Each split contains 24 layers, covering 216 full-data runs.}
\label{tab:phi-three-seed-layers}\\
\toprule
Layer & Forget ES $\downarrow$ & Retain ES $\uparrow$ & NFR $\uparrow$ & FQ $\uparrow$ & MU $\uparrow$ \\
\midrule
\endfirsthead
\caption[]{Full-data CoLU results for every Phi-1.5 layer (continued).}\\
\toprule
Layer & Forget ES $\downarrow$ & Retain ES $\uparrow$ & NFR $\uparrow$ & FQ $\uparrow$ & MU $\uparrow$ \\
\midrule
\endhead
\midrule
\multicolumn{6}{r}{\footnotesize Continued on next page.}\\
\endfoot
\bottomrule
\endlastfoot
\multicolumn{6}{l}{\textit{Forget-1\%}} \\
\midrule
0 & $0.0377\pm0.0099$ & $0.6012\pm0.0057$ & $91.78\pm0.83$ & $0.8916\pm0.1145$ & $0.4970\pm0.0035$ \\
\midrule[0.2pt]
1 & $0.0428\pm0.0111$ & $0.5829\pm0.0194$ & $90.02\pm2.31$ & $0.4743\pm0.1807$ & $0.5025\pm0.0067$ \\
\midrule[0.2pt]
2 & $0.0300\pm0.0058$ & $0.5977\pm0.0073$ & $92.14\pm0.97$ & $0.8169\pm0.0883$ & $0.5012\pm0.0035$ \\
\midrule[0.2pt]
3 & $0.0349\pm0.0008$ & $0.5785\pm0.0127$ & $90.32\pm1.00$ & $0.4626\pm0.1005$ & $0.5058\pm0.0054$ \\
\midrule[0.2pt]
4 & $0.0328\pm0.0028$ & $0.5896\pm0.0107$ & $91.31\pm1.00$ & $0.4163\pm0.1568$ & $0.5013\pm0.0034$ \\
\midrule[0.2pt]
5 & $0.0299\pm0.0054$ & $0.5678\pm0.0182$ & $89.92\pm0.96$ & $0.4046\pm0.0000$ & $0.5039\pm0.0025$ \\
\midrule[0.2pt]
6 & $0.0418\pm0.0043$ & $0.5815\pm0.0138$ & $89.99\pm1.00$ & $0.1054\pm0.0559$ & $0.5052\pm0.0043$ \\
\midrule[0.2pt]
7 & $0.0377\pm0.0043$ & $0.5718\pm0.0166$ & $89.60\pm1.42$ & $0.0828\pm0.0248$ & $0.5058\pm0.0014$ \\
\midrule[0.2pt]
8 & $0.0374\pm0.0037$ & $0.5528\pm0.0138$ & $88.20\pm0.87$ & $0.0828\pm0.0248$ & $0.5013\pm0.0028$ \\
\midrule[0.2pt]
9 & $0.0353\pm0.0042$ & $0.5464\pm0.0062$ & $87.89\pm0.38$ & $0.0456\pm0.0147$ & $0.4981\pm0.0027$ \\
\midrule[0.2pt]
10 & $0.0289\pm0.0021$ & $0.5552\pm0.0061$ & $89.06\pm0.41$ & $0.1985\pm0.0581$ & $0.5032\pm0.0068$ \\
\midrule[0.2pt]
11 & $0.0297\pm0.0033$ & $0.5510\pm0.0123$ & $88.69\pm1.07$ & $0.1197\pm0.0392$ & $0.4978\pm0.0046$ \\
\midrule[0.2pt]
12 & $0.0274\pm0.0035$ & $0.5567\pm0.0073$ & $89.29\pm0.73$ & $0.0971\pm0.0000$ & $0.5071\pm0.0045$ \\
\midrule[0.2pt]
13 & $0.0240\pm0.0021$ & $0.5601\pm0.0156$ & $89.81\pm1.21$ & $0.1423\pm0.0392$ & $0.5050\pm0.0034$ \\
\midrule[0.2pt]
14 & $0.0250\pm0.0062$ & $0.5786\pm0.0143$ & $91.11\pm0.82$ & $0.1423\pm0.0392$ & $0.5062\pm0.0056$ \\
\midrule[0.2pt]
15 & $0.0253\pm0.0008$ & $0.5601\pm0.0139$ & $89.71\pm0.99$ & $0.1533\pm0.0973$ & $0.5041\pm0.0031$ \\
\midrule[0.2pt]
16 & $0.0200\pm0.0017$ & $0.5627\pm0.0054$ & $90.33\pm0.48$ & $0.2321\pm0.0581$ & $0.5060\pm0.0011$ \\
\midrule[0.2pt]
17 & $0.0209\pm0.0032$ & $0.5951\pm0.0159$ & $92.67\pm0.94$ & $0.3120\pm0.0802$ & $0.4986\pm0.0026$ \\
\midrule[0.2pt]
18 & $0.0291\pm0.0024$ & $0.6118\pm0.0124$ & $93.26\pm1.06$ & $0.2784\pm0.1203$ & $0.5030\pm0.0044$ \\
\midrule[0.2pt]
19 & $0.0328\pm0.0026$ & $0.6135\pm0.0129$ & $93.09\pm0.77$ & $0.2448\pm0.1383$ & $0.4969\pm0.0030$ \\
\midrule[0.2pt]
20 & $0.0337\pm0.0019$ & $0.6406\pm0.0048$ & $95.04\pm0.47$ & $0.2079\pm0.1791$ & $0.5035\pm0.0021$ \\
\midrule[0.2pt]
21 & $0.0424\pm0.0050$ & $0.6165\pm0.0079$ & $92.55\pm0.94$ & $0.2222\pm0.1615$ & $0.5036\pm0.0010$ \\
\midrule[0.2pt]
22 & $0.0476\pm0.0044$ & $0.5881\pm0.0175$ & $90.01\pm1.20$ & $0.7544\pm0.1704$ & $0.4901\pm0.0036$ \\
\midrule[0.2pt]
23 & $0.0804\pm0.0099$ & $0.5988\pm0.0075$ & $88.19\pm0.39$ & $0.1197\pm0.0392$ & $0.4972\pm0.0042$ \\
\newpage
\multicolumn{6}{l}{\textit{Forget-5\%}} \\
\midrule
0 & $0.0073\pm0.0047$ & $0.5340\pm0.0222$ & $89.26\pm1.50$ & $(5.52\pm5.94)\!\times\!10^{-5}$ & $0.4835\pm0.0062$ \\
\midrule[0.2pt]
1 & $0.0129\pm0.0076$ & $0.5572\pm0.0193$ & $90.59\pm1.50$ & $0.0494\pm0.0803$ & $0.5033\pm0.0057$ \\
\midrule[0.2pt]
2 & $0.0046\pm0.0059$ & $0.5494\pm0.0066$ & $90.61\pm0.84$ & $0.0054\pm0.0094$ & $0.4997\pm0.0051$ \\
\midrule[0.2pt]
3 & $0.0131\pm0.0056$ & $0.5531\pm0.0176$ & $90.27\pm1.12$ & $0.1015\pm0.1469$ & $0.4957\pm0.0073$ \\
\midrule[0.2pt]
4 & $0.0180\pm0.0046$ & $0.5720\pm0.0169$ & $91.33\pm0.96$ & $0.3116\pm0.4797$ & $0.5038\pm0.0041$ \\
\midrule[0.2pt]
5 & $0.0221\pm0.0055$ & $0.5456\pm0.0145$ & $89.07\pm1.07$ & $0.0095\pm0.0059$ & $0.4926\pm0.0074$ \\
\midrule[0.2pt]
6 & $0.0236\pm0.0027$ & $0.5374\pm0.0041$ & $88.35\pm0.49$ & $0.0017\pm0.0012$ & $0.4997\pm0.0040$ \\
\midrule[0.2pt]
7 & $0.0194\pm0.0033$ & $0.5301\pm0.0152$ & $88.10\pm1.28$ & $0.0015\pm0.0014$ & $0.4981\pm0.0051$ \\
\midrule[0.2pt]
8 & $0.0216\pm0.0048$ & $0.5296\pm0.0125$ & $87.91\pm1.15$ & $(3.39\pm2.68)\!\times\!10^{-4}$ & $0.4956\pm0.0051$ \\
\midrule[0.2pt]
9 & $0.0230\pm0.0032$ & $0.5226\pm0.0169$ & $87.29\pm1.43$ & $(1.02\pm0.721)\!\times\!10^{-4}$ & $0.4938\pm0.0038$ \\
\midrule[0.2pt]
10 & $0.0214\pm0.0039$ & $0.5186\pm0.0099$ & $87.10\pm0.63$ & $(9.36\pm10.3)\!\times\!10^{-4}$ & $0.4941\pm0.0052$ \\
\midrule[0.2pt]
11 & $0.0164\pm0.0041$ & $0.5127\pm0.0081$ & $87.03\pm0.31$ & $0.0026\pm0.0015$ & $0.4863\pm0.0007$ \\
\midrule[0.2pt]
12 & $0.0186\pm0.0014$ & $0.5205\pm0.0084$ & $87.45\pm0.66$ & $(4.21\pm2.33)\!\times\!10^{-4}$ & $0.4829\pm0.0053$ \\
\midrule[0.2pt]
13 & $0.0240\pm0.0023$ & $0.5135\pm0.0140$ & $86.55\pm1.13$ & $(6.1\pm3.09)\!\times\!10^{-4}$ & $0.4907\pm0.0126$ \\
\midrule[0.2pt]
14 & $0.0265\pm0.0024$ & $0.5349\pm0.0026$ & $87.96\pm0.12$ & $(3.64\pm3.24)\!\times\!10^{-4}$ & $0.4856\pm0.0082$ \\
\midrule[0.2pt]
15 & $0.0307\pm0.0028$ & $0.5410\pm0.0087$ & $88.11\pm0.53$ & $(1.73\pm0.953)\!\times\!10^{-4}$ & $0.4912\pm0.0020$ \\
\midrule[0.2pt]
16 & $0.0233\pm0.0037$ & $0.5527\pm0.0109$ & $89.51\pm0.72$ & $0.0051\pm0.0060$ & $0.4945\pm0.0030$ \\
\midrule[0.2pt]
17 & $0.0445\pm0.0065$ & $0.5295\pm0.0060$ & $86.27\pm0.55$ & $0.0056\pm0.0033$ & $0.4955\pm0.0058$ \\
\midrule[0.2pt]
18 & $0.0203\pm0.0010$ & $0.5441\pm0.0013$ & $89.09\pm0.04$ & $0.1452\pm0.1590$ & $0.5091\pm0.0032$ \\
\midrule[0.2pt]
19 & $0.0417\pm0.0101$ & $0.5338\pm0.0288$ & $86.78\pm1.42$ & $0.0267\pm0.0221$ & $0.4967\pm0.0041$ \\
\midrule[0.2pt]
20 & $0.0242\pm0.0035$ & $0.5434\pm0.0168$ & $88.75\pm1.14$ & $0.0705\pm0.0622$ & $0.4965\pm0.0007$ \\
\midrule[0.2pt]
21 & $0.0604\pm0.0147$ & $0.5394\pm0.0380$ & $85.87\pm3.73$ & $0.0212\pm0.0147$ & $0.4940\pm0.0025$ \\
\midrule[0.2pt]
22 & $0.0686\pm0.0021$ & $0.5669\pm0.0172$ & $87.33\pm1.39$ & $0.0032\pm0.0025$ & $0.4910\pm0.0095$ \\
\midrule[0.2pt]
23 & $0.0090\pm0.0010$ & $0.5513\pm0.0021$ & $90.43\pm0.11$ & $0.0220\pm0.0176$ & $0.4745\pm0.0051$ \\
\newpage
\multicolumn{6}{l}{\textit{Forget-10\%}} \\
\midrule
0 & $0.0156\pm0.0125$ & $0.5284\pm0.0228$ & $88.26\pm2.17$ & $0.0489\pm0.0846$ & $0.4951\pm0.0108$ \\
\midrule[0.2pt]
1 & $0.0108\pm0.0117$ & $0.5447\pm0.0326$ & $89.82\pm2.79$ & $0.0010\pm0.0018$ & $0.4972\pm0.0041$ \\
\midrule[0.2pt]
2 & $0.0023\pm0.0022$ & $0.5614\pm0.0017$ & $91.66\pm0.16$ & $(4.96\pm7.85)\!\times\!10^{-7}$ & $0.4977\pm0.0071$ \\
\midrule[0.2pt]
3 & $0.0035\pm0.0015$ & $0.5632\pm0.0157$ & $91.71\pm1.22$ & $(6.77\pm5.66)\!\times\!10^{-9}$ & $0.4983\pm0.0083$ \\
\midrule[0.2pt]
4 & $0.0097\pm0.0070$ & $0.5115\pm0.0249$ & $87.42\pm2.18$ & $0.0069\pm0.0120$ & $0.5007\pm0.0104$ \\
\midrule[0.2pt]
5 & $0.0033\pm0.0026$ & $0.5327\pm0.0340$ & $89.45\pm2.70$ & $(2.95\pm5.1)\!\times\!10^{-8}$ & $0.4978\pm0.0048$ \\
\midrule[0.2pt]
6 & $0.0102\pm0.0028$ & $0.4998\pm0.0186$ & $86.51\pm1.57$ & $0.0219\pm0.0380$ & $0.5002\pm0.0061$ \\
\midrule[0.2pt]
7 & $0.0076\pm0.0027$ & $0.4670\pm0.0092$ & $84.25\pm0.78$ & $0.0011\pm0.0012$ & $0.4898\pm0.0086$ \\
\midrule[0.2pt]
8 & $0.0021\pm0.0015$ & $0.4456\pm0.0117$ & $83.04\pm0.96$ & $(5.72\pm6.29)\!\times\!10^{-4}$ & $0.4964\pm0.0050$ \\
\midrule[0.2pt]
9 & $0.0078\pm0.0008$ & $0.4859\pm0.0049$ & $85.65\pm0.38$ & $0.0080\pm0.0072$ & $0.4997\pm0.0066$ \\
\midrule[0.2pt]
10 & $0.0045\pm0.0020$ & $0.4885\pm0.0180$ & $86.08\pm1.31$ & $(1.45\pm1.28)\!\times\!10^{-5}$ & $0.5023\pm0.0032$ \\
\midrule[0.2pt]
11 & $0.0056\pm0.0016$ & $0.5238\pm0.0217$ & $88.63\pm1.51$ & $(2\pm2.7)\!\times\!10^{-6}$ & $0.4990\pm0.0073$ \\
\midrule[0.2pt]
12 & $0.0061\pm0.0014$ & $0.4956\pm0.0152$ & $86.49\pm1.19$ & $0.0195\pm0.0213$ & $0.4995\pm0.0046$ \\
\midrule[0.2pt]
13 & $0.0077\pm0.0020$ & $0.4692\pm0.0086$ & $84.41\pm0.55$ & $0.0016\pm0.0022$ & $0.4881\pm0.0028$ \\
\midrule[0.2pt]
14 & $0.0275\pm0.0088$ & $0.5098\pm0.0117$ & $86.03\pm0.77$ & $(2.67\pm1.75)\!\times\!10^{-4}$ & $0.4956\pm0.0058$ \\
\midrule[0.2pt]
15 & $0.0534\pm0.0018$ & $0.5361\pm0.0049$ & $86.15\pm0.24$ & $(7.8\pm3.92)\!\times\!10^{-5}$ & $0.5013\pm0.0008$ \\
\midrule[0.2pt]
16 & $0.0452\pm0.0085$ & $0.5392\pm0.0053$ & $86.96\pm0.25$ & $(2.69\pm0.566)\!\times\!10^{-4}$ & $0.4973\pm0.0036$ \\
\midrule[0.2pt]
17 & $0.0696\pm0.0074$ & $0.5491\pm0.0084$ & $85.97\pm0.33$ & $(8.29\pm4.61)\!\times\!10^{-4}$ & $0.5038\pm0.0026$ \\
\midrule[0.2pt]
18 & $0.0425\pm0.0061$ & $0.5505\pm0.0069$ & $88.00\pm0.95$ & $0.0042\pm0.0047$ & $0.5041\pm0.0028$ \\
\midrule[0.2pt]
19 & $0.0568\pm0.0121$ & $0.5750\pm0.0050$ & $88.81\pm0.93$ & $(7.79\pm8.14)\!\times\!10^{-4}$ & $0.4986\pm0.0017$ \\
\midrule[0.2pt]
20 & $0.0544\pm0.0111$ & $0.5618\pm0.0073$ & $88.00\pm0.53$ & $(5.25\pm3.3)\!\times\!10^{-4}$ & $0.5009\pm0.0022$ \\
\midrule[0.2pt]
21 & $0.0719\pm0.0253$ & $0.5258\pm0.0117$ & $84.07\pm2.66$ & $0.0029\pm0.0038$ & $0.4910\pm0.0058$ \\
\midrule[0.2pt]
22 & $0.0601\pm0.0068$ & $0.5446\pm0.0071$ & $86.30\pm0.84$ & $(4.92\pm3.62)\!\times\!10^{-4}$ & $0.4952\pm0.0011$ \\
\midrule[0.2pt]
23 & $0.0079\pm0.0044$ & $0.5205\pm0.0085$ & $88.21\pm0.33$ & $(4.46\pm6.93)\!\times\!10^{-4}$ & $0.4785\pm0.0097$ \\
\end{longtable}
\endgroup

\end{document}